\documentclass[journal]{IEEEtran}
\ifCLASSINFOpdf
\else
\fi
\usepackage{amsmath}
\usepackage{algorithmic,algorithm}

\usepackage{float}
\usepackage{array}
\usepackage{multirow}
\usepackage[table,xcdraw]{xcolor}
\usepackage{booktabs}

\usepackage{mathrsfs}

\ifCLASSOPTIONcompsoc
  \usepackage[caption=false,font=normalsize,labelfont=sf,textfont=sf]{subfig}
\else
  \usepackage[caption=false,font=footnotesize]{subfig}
\fi
\usepackage{graphicx}

\makeatletter
\def\endthebibliography{%
	\def\@noitemerr{\@latex@warning{Empty `thebibliography' environment}}%
	\endlist
}
\makeatother

\begin{document}
%
\title{TBC-Net: A real-time detector for infrared small target detection using semantic constraint}
%
%
%

\author{Mingxin Zhao,~\IEEEmembership{Student Member,~IEEE},
		Li Cheng,
		Xu Yang,
		Peng Feng,
		Liyuan Liu,~\IEEEmembership{Member,~IEEE}, \\ and 
		Nanjian Wu,~\IEEEmembership{Member,~IEEE}
\thanks{This work was supported by National Natural Science Foundation of China
	(Grant No. 61704167, 61434004), Beijing Municipal Science and Technology Project
	(Z181100008918009), Youth Innovation Promotion Association Program, Chinese
	Academy of Sciences (No.2016107), the Strategic Priority Research Program of Chinese
	Academy of Science, Grant No.XDB32050200. (Corresponding author: Nanjian Wu.)
	}
\thanks{The authors are with the State Key Laboratory of Superlattices and Microstructures, Institute of Semiconductors, Chinese Academy of Sciences, Beijing 100083, China, and also with the Center of Materials Science and Optoelectronics Engineering, University of Chinese Academy of Sciences, Beijing 100049, China, and also with the Center for Excellence in Brain Science and Intelligence Technology, Chinese Academy of Sciences, Beijing 100083, China (email: zhaomingxin17@semi.ac.cn; chengli17@semi.ac.cn; yangxu@semi.ac.cn; fengpeng06@semi.ac.cn; liuly@semi.ac.cn; nanjian@red.semi.ac.cn).}
}

%
%

\markboth{Journal of \LaTeX\ Class Files,~Vol.~14, No.~8, August~2015}%
{Shell \MakeLowercase{\textit{et al.}}: Bare Demo of IEEEtran.cls for IEEE Journals}
%



\maketitle

\begin{abstract}

Infrared small target detection is a key technique in infrared search and tracking (\textbf{IRST}) systems. Although deep learning has been widely used in the vision tasks of visible light images recently, it is rarely used in infrared small target detection due to the difficulty in learning small target features. In this paper, we propose a novel lightweight convolutional neural network TBC-Net for infrared small target detection. The TBC-Net consists of a target extraction module (TEM) and a semantic constraint module (SCM), which are used to extract small targets from infrared images and to classify the extracted target images during the training, respectively. Meanwhile, we propose a joint loss function and a training method. The SCM imposes a semantic constraint on TEM by combining the high-level classification task and solve the problem of the difficulty to learn features caused by class imbalance problem. During the training, the targets are extracted from the input image and then be classified by SCM. During the inference, only the TEM is used to detect the small targets. We also propose the data synthesis method to generate the training data. The experimental results show that compared with the traditional methods, TBC-Net can better reduce the false alarm caused by complicated background, the proposed network structure and joint loss have a significant improvement on small target feature learning. Besides, TBC-Net can achieve real-time detection on the NVIDIA Jetson AGX Xavier development board, which is suitable for applications such as field research with drones equipped with infrared sensors.

\end{abstract}

\begin{IEEEkeywords}
Convolutional neural network, infrared images, small target detection, semantic feature
\end{IEEEkeywords}

%
\IEEEpeerreviewmaketitle

\section{Introduction}
\begin{figure}[!h]
	\centering
	\subfloat[Tracking wildlife]{\includegraphics[width=1.1in]{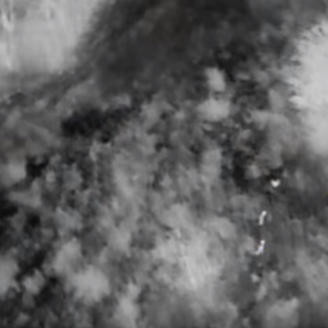}%
	\label{wolf}}
	\hfil
	\subfloat[Field search and rescue]{\includegraphics[width=1.1in]{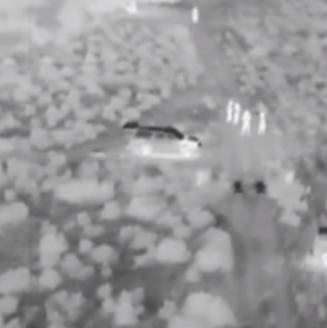}%
	\label{people}}
	\centering
	\caption{Use infrared sensor on drones for scientific research and field rescue}
	\label{drone}
\end{figure}

%
%
%
%
\IEEEPARstart{I}{nfrared} small target detection is a key technique in infrared search and tracking (IRST) systems. IRST has excellent potential in scientific research and civil applications. With the rise and widespread use of drones in recent years, infrared sensors are increasingly being carried by drones to perform field detection, rescue, and precise positioning \cite{rudol2008human}\cite{karma2015use}\cite{silvagni2017multipurpose} (as shown in Figure \ref{drone}). However, limited by the resolution of the infrared sensor, atmospheric scattering, background temperature noise, etc., the imaging quality of the infrared image is generally worse than that of the visible light sensor. Moreover, in most of these applications, the target size in an infrared image is often less than 10 pixels. Therefore, there are more and more application prospects and technical challenges for the detection of small infrared targets.

In order to solve the problem of small target detection, many methods have been proposed. These methods can generally be classified into two categories: single-frame detection and multi-frame detection. Since multi-frame detection algorithms usually consume more time than single-frame detection algorithms, and such algorithms generally assume that the background is static \cite{qin2019infrared}, which make the multi-frame detection algorithms unsuitable for unmanned aerial vehicle (UAV) applications. In this paper, we focus on the single-frame detection algorithms. 

The traditional single-frame detection methods based on morphological filtering consider that the small target belongs to the high-frequency components in the image so that the small target can be separated from the background by the filtering. For example, 2-D least-mean-square (TDLMS) adaptive filtering \cite{soni1993performance}, Max-Mean/Max-Median filtering \cite{rivest1996detection}, and Top-Hat filtering \cite{bai2010analysis} are all such methods. These methods are easily affected by the clutters and noise present in the background, which affects the robustness of the detection. 

In recent years, the small target detection methods based on the human visual system (HVS) are mainly used to distinguish the target from the background by constructing different local contrast measures. The local contrast measure (LCM) \cite{chen2013local}, the improved LCM (ILCM) \cite{han2014robust}, novel local contrast measure (NLCM) \cite{qin2016effective}, weighted new local contrast measure (WLCM) \cite{liu2018tiny}, novel weighted image entropy map (NWIE) \cite{deng2016infrared}, multiscale patch-based contrast measure (MPCM) \cite{wei2016multiscale}, high-boost-based multiscale local contrast measure (HB-MLCM) \cite{shi2017high} are all HVS-based methods. These methods construct an internal window and its adjacent windows, or surrounding areas, in a local area to calculate the contrast between the internal window and adjacent windows or surrounding areas to enhance the local target features. The detection of the target is achieved by sliding the internal window over the entire image and finally using adaptive threshold segmentation. However, these algorithms are also susceptible to factors such as edges and noise.

Due to the great success of deep neural networks in natural image processing, some works \cite{fan2018dim, liangkui2018using, wang2017small} have begun to introduce deep learning, especially convolutional neural networks (CNN), into infrared small target detection. However, most CNN algorithms do not perform well in learning small target features \cite{gao2018robust} and take a long time to run inference. For example, Mask-RCNN \cite{he2017mask} takes a one-fifth second to perform an image detection on the GPU. Lightweight networks such as YOLO \cite{redmon2016you} and SSD \cite{liu2016ssd} are compact and fast, but the detection is less effective. Moreover, the above algorithms always perform poorly on small objects. In the practice of CNN methods in infrared small target detection, Fan \textit{et al.} \cite{fan2018dim} used the MNIST dataset to train a multiscale CNN and extracted the convolution kernels to enhance small target images. Lin \textit{et al.} \cite{liangkui2018using} designed a seven-layer network to detect small targets by learning the synthesis data generated by oversampling. Wang \textit{et al.} \cite{wang2017small} transferred a CNN pre-trained on ILSVRC 2013 to small target datasets to learn small target features. However, objects in the real world usually contain a large number of shapes, colors, and structural information, which are not available in small targets. The effectiveness of transfer learning is limited.
%

In the field application scenarios of drones, the background often has a lot of clutters such as branches, roads, buildings. Compared with the sky, clouds, and sea surface, the background composition is more complicated. Intuitively, if a broader range of image features can be utilized, it should be helpful to suppress these complicated interferences and reduce the false alarm rate, which is difficult to achieve by traditional methods based on local features.

\begin{figure}[!h]
	\centering
	\includegraphics[width=0.38\textwidth]{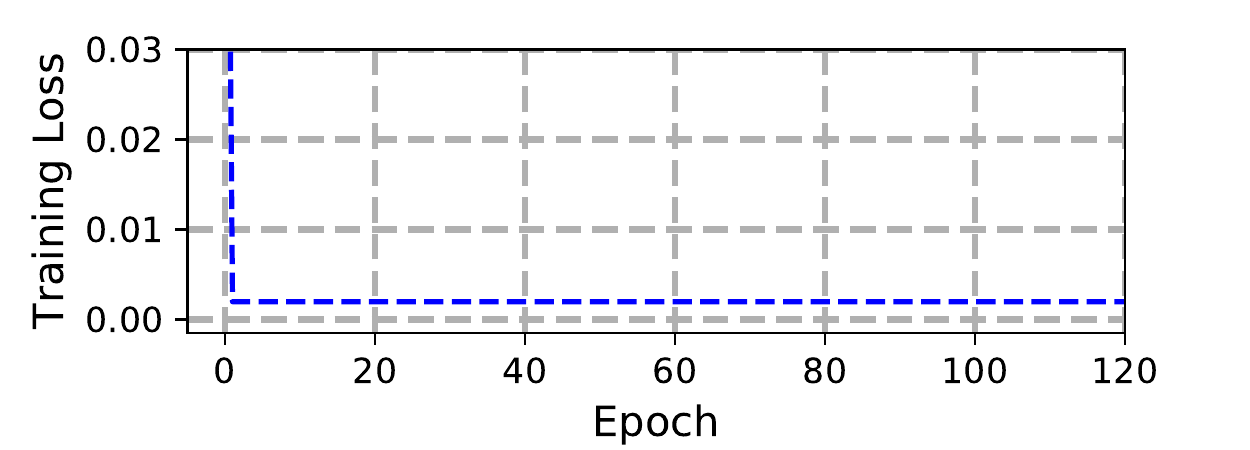}
	\caption{A training loss curve caused by the class imbalance problem.}
	\label{loss_example}
\end{figure}

Encouraged by the progress of image segmentation networks in recent years, we hope to segment the target image from the original image directly. However, using a natural image segmentation network to segment a small target can cause some problems. First of all, the artifact problem seriously degrade segmentation and detection performance. Secondly, the small target occupies a small proportion in the entire image, so that the training process encounters severe class imbalance problem \cite{xue2018segan}. In actual training, the convergence curve shown in Figure \ref{loss_example} is prone to appear, and the network converges quickly, but the small target features cannot be adequately learned. 

Although using a small image of $28\times 28$ as used in \cite{liangkui2018using} can alleviate the class imbalance problem, the use of small images limits the network to learn the interference characteristics of the background from a broader range, and it is impossible to make full use of the deeper receptive field of the network to suppress background interference.

To solve the above problems, we propose a novel segmentation convolutional neural network called TBC-Net and design the corresponding training method. Compared with traditional methods and HVS-based methods, the proposed network can achieve better detection performance. The contributions of this paper are as follows:

\begin{enumerate}
	\item A lightweight infrared small target detection neural network called TBC-Net is proposed, which includes a target extraction module and a semantic constraint module.
	\item A novel training method is proposed to solve the problem of extreme imbalance of foreground and background in small target image by adding high-level semantic constraint information of images in training.
	\item It can achieve real-time detection of $256\times 256$ images on the NVIDIA Jetson AGX Xavier embedded development board.
\end{enumerate}

The remainder of this paper is organized as follows: In Section II, we briefly review the technical background related to TBC-Net. In Section III, we introduce the network structure and detection method and analyze the storage and computational complexity of the network. In Section IV, we introduce loss function design and training methods. In Section V, we experiment with real infrared sequences. Finally, we present the conclusions of this paper in Section VI.

\section{Background}

In this section, we briefly introduce the technical background of TBC-Net, including CNN-based image segmentation, residual learning, and semantic constraint, so that readers can better understand the design ideas of TBC-Net in subsequent sections.

\subsection{CNN-based Segmantation}

The fully convolutional neural network (FCN) \cite{long2015fully} was proposed to use CNN for image segmentation by replacing the fully connected layer with the convolutional layer. However, FCN uses VGG-Net \cite{simonyan2014very} as its feature extraction network and takes one-fifth of a second to complete an image segmentation, which is difficult to achieve real-time performance. Noh \textit{et al.} \cite{noh2015learning} used deconvolution to upsample the feature map and obtain the segmentation mask. There are also many works such as \cite{mohan2014deep}\cite{saito2016real} based on deconvolution to achieve image segmentation. Nevertheless, there are also some artifacts in images produced by deconvolution.

Image segmentation can also be seen as a pixel-wise classification task. Chen \textit{et al.} \cite{chen2017deeplab} used a convolution operation called "dilated convolution" to achieve pixel-wise classification. Yu \textit{et al.} \cite{yu2017dilated} proposed dilated residual networks (DRN) to reduce the checkerboard artifacts caused by dilated convolution.

In order to fuse the multiscale information, Lin \textit{et al.} \cite{lin2016efficient} used multiscale image input and sliding pyramid pooling to improve the performance. Deeplab \cite{chen2017deeplab} used multiscale input images to extract features. U-Net \cite{ronneberger2015u} realized multiscale feature fusion by concatenating the feature maps obtained by downsampling with the upsampled feature maps of the same scale.

All the above algorithms are aimed at natural images; for small infrared targets that lack complicated features within 10 pixels, there is still no effective deep-learning-based segmentation solution.

\subsection{Residual Learning}

ResNet \cite{he2016deep} was proposed to use residual connections to enhance the learning effect of CNN. Highway Networks \cite{srivastava2015highway} and DenseNet \cite{huang2017densely} have proven that residual connection is effective both in improving convergence property and enhancing network performance. Residual connections are short paths from early layers to later layers. On the one hand, these cross-layer connections can solve the problems that the network is difficult to converge, and the inference accuracy is reduced due to the vanishing-gradient problem, etc.. The shallow information of the network can be directly transmitted to the deeper layer so that the network can more effectively learn the image features contained in the shallow layer. DnCNN \cite{zhang2017beyond} used the idea of residual learning to construct a network for image denoising, which has achieved good noise reduction visual effects.

\subsection{Semantic Constraint}

Liu \textit{et al.} \cite{liu2017image} proposed the idea of using a high-level vision task to enhance a denoise neural network. They cascaded a denoise CNN with various high-level vision tasks and used the joint loss to update the weights of the denoise CNN during training, thereby improving the visual effect of the output image. This training method, combined with image semantic constraint, has a significant improvement on learning image features such as noise that are difficult to describe.

Following this idea, Wang \textit{et al.} \cite{wang2019segmentation} adopted image denoising as a low-level vision task and image segmentation as a high-level vision task. Then they trained the joint pipeline using hybrid losses to improve denoising effect.

\section{Proposed Network: TBC-Net}

In general, an infrared small target image can be expressed by the following formula \cite{gu2010kernel}: 
\begin{equation}
f_D(x,y)=f_T(x,y)+f_B(x,y)+f_N(x,y)
\label{basic_formula}
\end{equation}
where $f_D$, $f_T$, $f_B$, $f_N$ and $(x,y)$ are original infrared image, the small targets image, the background, the noise and the pixel location, respectively. In the following parts, we omit the $(x, y)$ without causing confusion.

\subsection{Network Architecture}

\begin{figure*}[!h]
	\centering
	\hfil
	\includegraphics[width=0.58\textwidth]{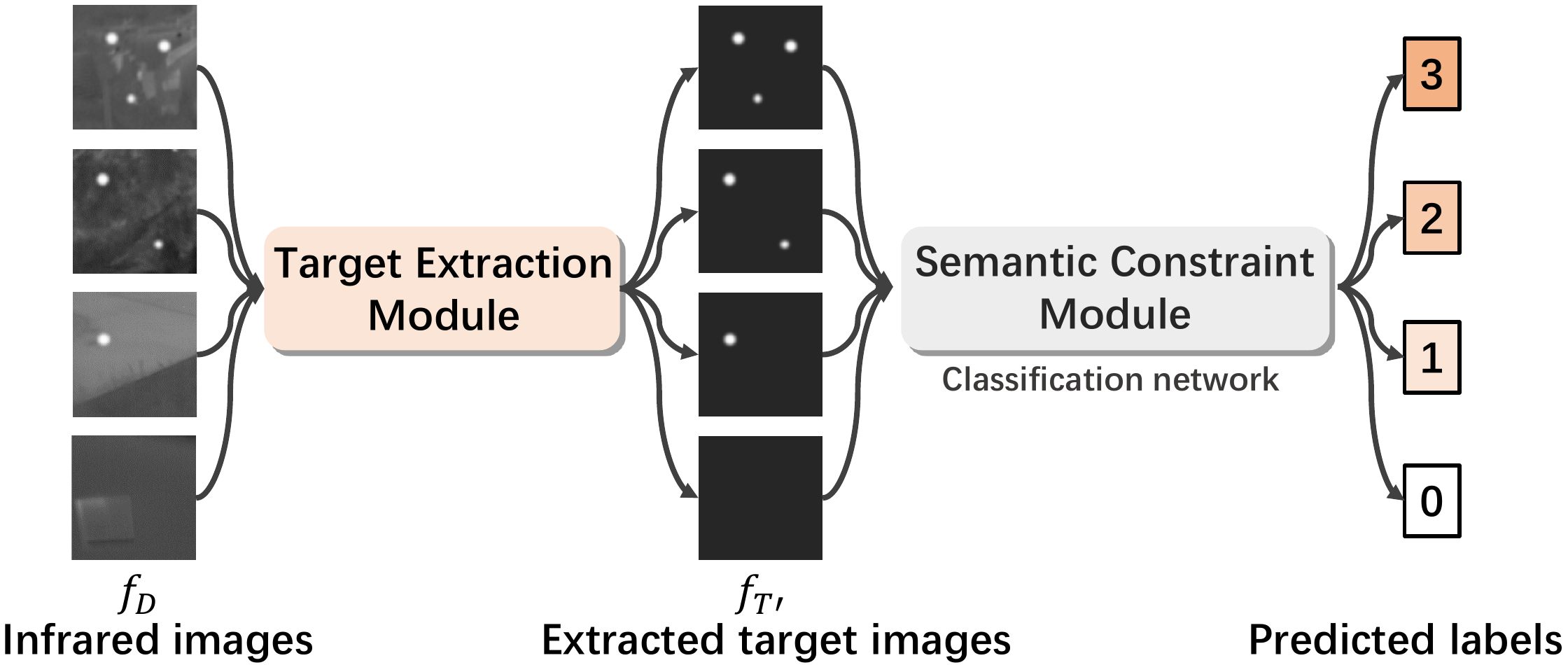}
	\caption{The architecture of TBC-Net.}
	\label{semantic_use}
\end{figure*}

Our proposed network, as shown in Figure \ref{semantic_use}, consists of two modules: target extraction module (TEM) and semantic constraint module (SCM). We name the network TBC-Net, and the reason for the name will be explained at the beginning of Section IV. The TEM is a lightweight image segmentation network with compact operations and flexible structure parameters for efficient inference. The SCM is a multi-layer CNN used to achieve high-level classification task.

The input infrared image is processed by TEM to obtain the target image $f_T$. The SCM classifies the target image $f_T$ according to the number of the targets contained in $f_D$. This high-level vision task can add semantic constraint into the TBC-Net and improve the TEM performance during the training phase. When training TBC-Net, the SCM needs to be pre-trained on the synthesis data, and then when training the TEM, the SCM parameters remain unchanged. The quality of $f_{T}$ obtained by TEM will have an impact on the classification results of SCM, and then the constraint information brought by image semantics will be transferred to TEM through back-propagation. The existence of SCM solves the problem that it is difficult to learn the features caused by the imbalance between the target and the background in the small target data, so that the compact TEM can still effectively learn the small target features. To train the network, we propose a joint loss function and a corresponding training method, which are shown in Section IV.

During the inference phase, only the TEM plays a role in extracting targets. Therefore, in practical applications, the inference speed of the network depends only on the complexity of the TEM.

Below we introduce the design ideas and details of TEM and SCM.

\subsection{Target Extraction Module}

Compared with natural image datasets such as ImageNet \cite{deng2009imagenet}, Pascal VOC \cite{everingham2010pascal}, and MS COCO \cite{lin2014microsoft}, the infrared small target image does not contain color information. Meanwhile, because the target size is small, it does not contain category, and shape information. Networks used in natural image semantic and instance segmentation generally need to learn the color, shape and category features. However, such problems do not exist on infrared small target data. Therefore, to improve the network's inference efficiency, we design a compact TEM module with a more lightweight network structure by using the characteristics of the small infrared target data.

\begin{figure*}[!h]
	\centering
	\includegraphics[width=0.8\textwidth]{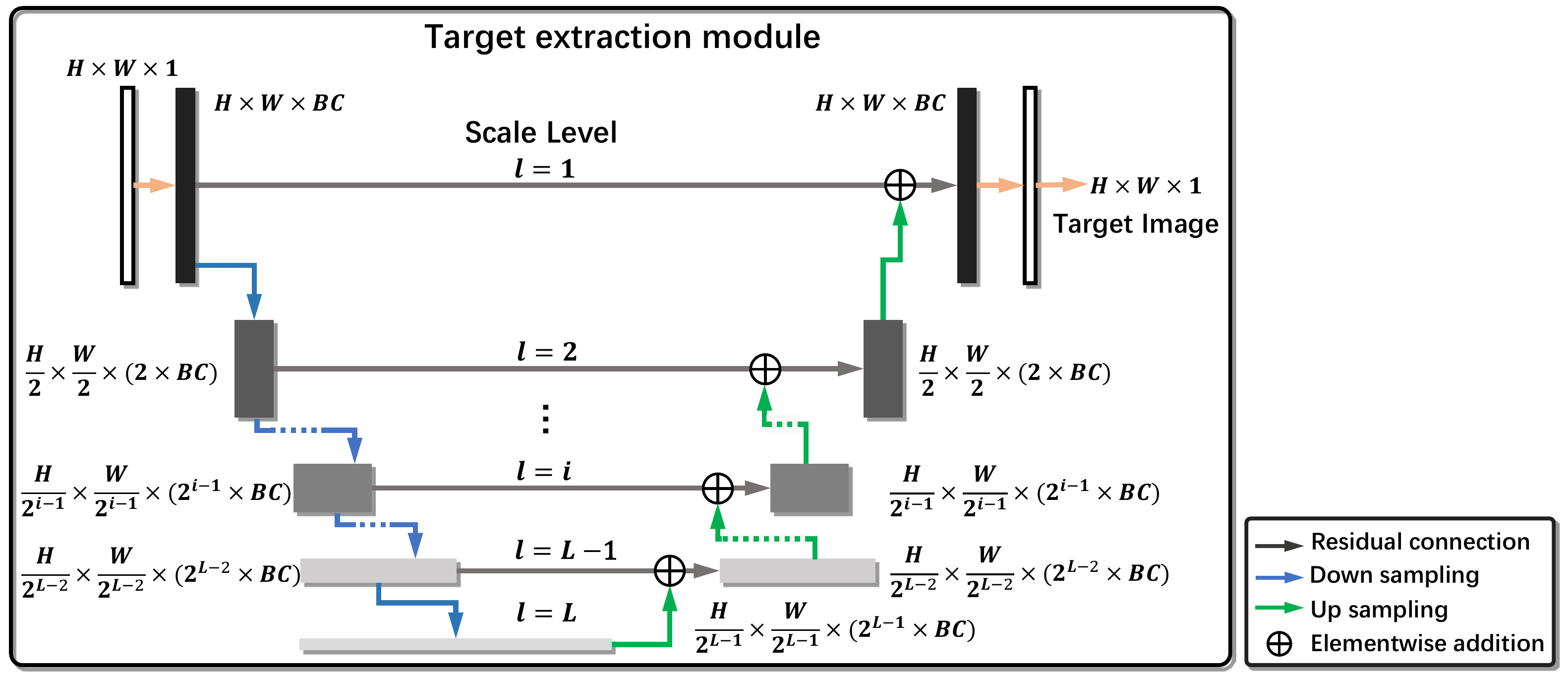}
	\caption{Target extraction module of TBC-Net}
	\label{TEM}
\end{figure*}
\begin{figure}[!h]
	\centering
	\subfloat[Downsample module]{\includegraphics[height=0.14\textwidth]{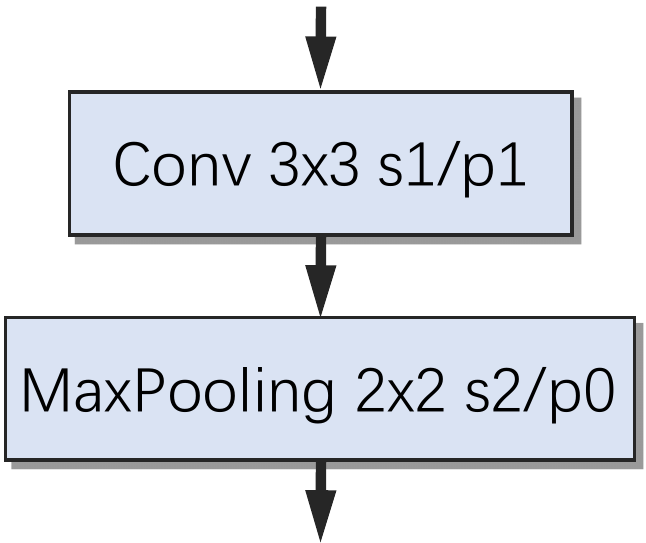}%
		\label{downsample}}
	\hfil
	\subfloat[Upsample module]{\includegraphics[height=0.14\textwidth]{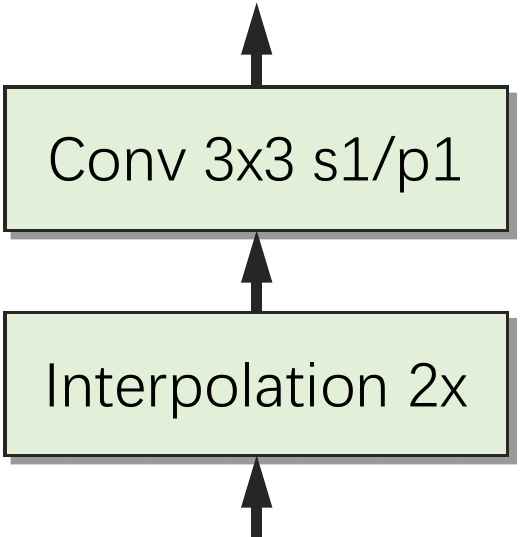}%
		\label{upsample}}
	\hfil
	\caption{Downsample module and upsample module used in TBC-Net}
\end{figure}

Based on the above analysis, we use compact operations to implement the downsampling and upsampling modules to form the "Encoder-Decoder" structure commonly used in the image segmentation field. The structure of the TEM is shown in Figure \ref{TEM}. We use the 2D convolutional layer and the MaxPooling layer to form the downsampling module (as shown in Figure \ref{downsample}) and refer to \cite{odena2016deconvolution} to use the nearest neighbor interpolation and 2D convolutional layer to form the upsampling module (as shown in Figure \ref{upsample}). The upsampling features are fused with the downsampling features of the same scale by the residual connections. The upsampling modules, the downsampling modules, and the residual connections together form the TEM.

The TEM is used to extract target image $f_{T'}$, and the formal expression is as follows:
\begin{equation}
\begin{aligned}
f_{T'}&=TEM(f_D) \\
\end{aligned}
\label{recons_bkg}
\end{equation}
where $f_{T'}$ denotes the target image output by TEM.

The infrared image is generally a single-channel grayscale image. The number of input channels is firstly expanded using a 2D convolution layer. The number of channels that the input layer expands is called base channels (\textbf{$BC$}). The number of downsampling operations determines how many scales the network needs to fuse, so it is also a critical network structure parameter. We name the maximum scale level produced by downsampling $L$. By changing $BC$ and $L$, the structure of the network can be adjusted accordingly. Besides, $BC$ and $L$ also affect the parameter storage space and the amount of computation of the algorithm.



\begin{figure}[!h]
\centering
\subfloat[Interference edge]{\includegraphics[width=1in]{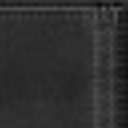}%
\label{zeropadding}}
\hfil
\subfloat[Checkerboard artifacts on small targets]{\includegraphics[width=1in]{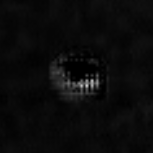}%
\label{checkerboard}}
\hfil
\caption{Problems caused by zero padding and deconvolution.}
\label{problems}
\end{figure}

It is worth noting that we do not use zero paddings to avoid interference edges of the target image, and do not use deconvolution to avoid checkerboard artifacts. Both of these affect the detection of small targets (as shown in Figure \ref{problems}).

\subsection{Semantic Constraint Module}

Although in edge segmentation, vessel segmentation, and other applications, there have been many methods such as training on small patches, using weighted loss functions to solve the sample class imbalance problem. However, as mentioned above, training on small size images cannot enable the network to learn the complex interference information of the background, and cannot effectively utilize the large receptive field of the network. Meanwhile, the choice of weights in loss function is task-specific and is hard to optimize \cite{xue2018segan}.

Motivated by \cite{liu2017image}, we believe that adding semantic information during the training can enable the network to learn the features of small targets better. Nevertheless, small infrared targets often have only 2 to 10 pixels, and there are a large number of non-ideal factors in the infrared imaging process, making it difficult to describe them with high-level semantics such as shapes and categories. According to the characteristics of the small target image, we propose a direct and straightforward image semantic constraint, that is, counting the number of small targets in the target image. The method of using this semantic information is as follows:

\begin{enumerate}
	\item Use TEM to extract target image $f_{T'}$
	\item Use another network to predict the number of targets contained in $f_{T'}$
\end{enumerate}

We use a CNN to classify $f_{T'}$, and its structure is shown in Table \ref{scb}, where $C_{SCM}$ is the number of classes corresponding to the fully connected layer. We call the classification network as the semantic constraint module to illustrate its guiding role at the semantic level during TEM training. This process is also illustrated in Figure \ref{semantic_use}.

\begin{table}[!h]
	\renewcommand{\arraystretch}{1.3}
	\centering
	\caption{Network structure of semantic constraint module used in TBC-Net ($H=256,W=256$)}
	\label{scb}
	\begin{tabular}{|l|l|l|}
		\hline
		\textbf{Type / Stride / Padding} & \textbf{Filter Shape} & \textbf{Input Size} \\ \hline
		Conv / s1 / p1                   & $3\times 3\times 1\times 32$                       & $256\times 256\times 1$           \\ \hline
		MaxPooling / s2 / p0             & $2\times 2$                            & $256\times 256\times 32$          \\ \hline
		Conv / s1 / p1                   &  $3\times 3\times 32\times 64$                      & $128\times 128\times 32$        \\ \hline
		MaxPooling / s2 / p0             & $2\times 2$                             & $128\times 128\times 64$           \\ \hline
		Conv / s1 / p1                   & $3\times 3\times 64\times 32$                     & $64\times 64\times 64$            \\ \hline
		MaxPooling / s2 / p0             & $2\times 2$                            & $64\times 64\times 32$           \\ \hline
		Conv / s1 / p1                   & $3\times 3\times 32\times 16$                      & $32\times 32\times 32$           \\ \hline
		MaxPooling / s2 / p0             & $2\times 2$                            & $32\times 32\times 16$            \\ \hline
		AvgPooling / s4 / p0             & $4\times 4$                            & $16\times 16\times 16$         \\ \hline
		FC                               & $256\times C_{SCM}$             & $256$                 \\ \hline
	\end{tabular}
\end{table}

\subsection{Segmentation and Detection}

After the TEM obtains the target image, we use the adaptive threshold method on the target image to obtain the binarized segmentation image. The calculation method of the adaptive threshold is defined as follows:

\begin{equation}
T= \mu + k\times \sigma
\end{equation}
where $T$ is the segmentation threshold, $\mu$ and $\sigma$ are the mean and standard deviation of the TEM output image, respectively. $k$ is the empirical parameter which ranges from 20 to 30 in our experiment. 

\begin{figure*}[!h]
	\centering
	\includegraphics[width=0.8\textwidth]{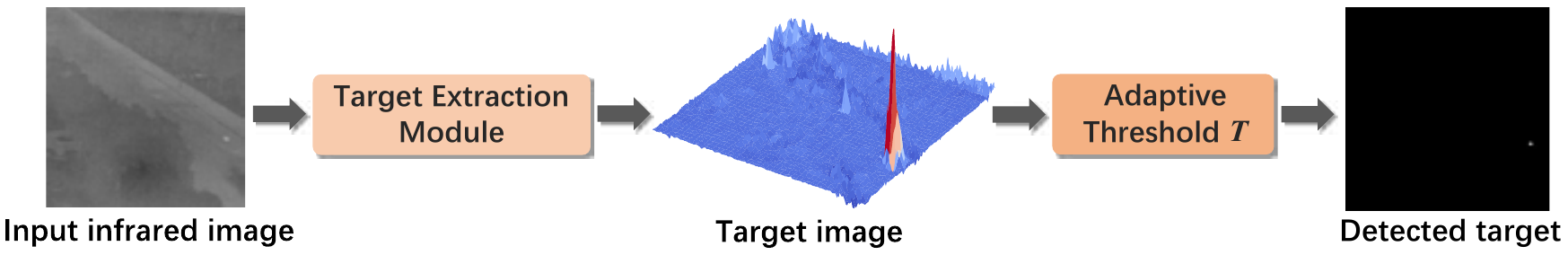}
	\caption{Using TBC-Net to detect infrared small targets.}
	\label{workflow}
\end{figure*}

The complete workflow of using TBC-Net for small target detection is shown in Figure \ref{workflow}.

\subsection{Storage and Computation Analysis}

Generally, only the computational complexity is analyzed in the algorithm analysis. However, for the practical application scenarios of infrared small target detection, the detection algorithm generally runs on hardware devices with limited storage space and computation resources, while the parameter storage requirements of the neural network are relatively large. So, in addition to the computational complexity analysis, we also analyze the parameter storage requirements.

\subsubsection{Computational Complexity Analysis}

In the computational complexity analysis, we do not use the $O$ symbol, but use $OPs$ (the number of multiply-add operations) commonly used in neural network design \cite{zhang2018shufflenet} as an indicator to measure the amount of calculation. The nearest neighbor interpolation used in upsampling modules does not require multiply and accumulate operations, so it is ignored in the calculation of $OPs$. The $OPs$ of the 2D convolution layer is as follows:

\begin{equation}
\begin{aligned}
OPs_{conv2d}=k_{c}^2\times h\times w \times \frac{C_{in}}{s} \times \frac{C_{out}}{s}
\end{aligned}
\end{equation}
where $h$ and $w$ are the height and width of the input feature map, $C_{in}$ and $C_{out}$ are the number of input and output channels, respectively, $k_c$ is the convolution kernel size, and $s$ is the stride size. In the 2d convolution layer of TBC-Net, $k_c=3$, $s=1$. Therefore, the $OPs$ of TBC-Net can be obtained by accumulating the $OPs$ of all the downsampling and upsampling modules.

\begin{equation}
\begin{aligned}
OPs&=9\sum_{l=1}^{L}{\frac{H}{2^{l-1}}\times \frac{W}{2^{l-1}}\times (BC\times 2^{l-1})\times (BC\times 2^l)} \\
&=\frac{9}{2} BC^2\times HW\times L
\end{aligned}
\end{equation}
where $H$ and $W$ are the height and width of the TBC-Net input image, respectively.

The calculation shows that the computational complexity of the algorithm is proportional to the number of pixels of the input image and the maximum scale level $L$, respectively, and is squared with the number of base channel $BC$.

\subsubsection{Parameter Storage Analysis}
The storage requirement consists of two parts, the feature map storage $M_m$ and the parameter storage $M_p$.

During the inference phase of TBC-Net, the previous feature maps cannot be discarded until the first upsampling. After the first upsampling, the downsampled feature map corresponding to the upsampling scale can be discarded. So the maximum storage requirement of feature maps $M_{m}$ during the inference occurs before the first upsampling:

\begin{equation}
\begin{aligned}
M_{m}&=H\times W\times(2 + 2\sum_{l=1}^{L-1}{\frac{BC}{2^{l-1}}} + \frac{BC}{2^{L-1}})\\ 
  &=[BC\times (4-\frac{6}{2^{L}})+2] M_{input} \\
\end{aligned}
\end{equation}
where $M_{input}$ is the number of pixels of the input image.

In terms of network weights, its storage space mainly depends on $BC$ and $L$. The number of parameters of a 2-D convolution layer is $k_{c}^2\times C_{in}\times C_{out}$, the parameter storage space is calculated as follows:

\begin{equation}
\begin{aligned}
M_{p}&=9\times \sum_{l=1}^{L}(2^{l-1} BC \times 2^{l} BC) \times 2 + 2\times 9\times BC\\
&= 12\times(4^L-1)\times BC^2 + 18 BC \\
\end{aligned}
\end{equation}

The first part is the amount of weights storage required for the downsampling and upsampling modules, and the second part is the weights storage of the input and output layer. It can be seen that the parameter storage space is approximately proportional to the square of $BC$ and exponential with $L$. Furthermore, we can get the total storage requirement $M_{TBC}$ of TBC-Net:
\begin{equation}
\begin{aligned}
M_{TBC}=M_{m}+M_{p}
\end{aligned}
\end{equation}

\begin{table}[!h]
\renewcommand{\arraystretch}{1.3}
\centering
\caption{Storage and computation under different $BC$ and $L$ ($H=256,W=256$)}
\label{net_config}
\begin{tabular}{c|c|c|c|c|c}
\hline
$BC$ & $L$ & $M_{m}$ (M)    & $M_{p}$ (M)    & $M_{TBC}$ (M)    & $OPs$ (M) \\ \hline
8  & 3 & 1.750  & 0.046 & 1.796 & 54   \\ \hline
8  & 4 & 1.938     & 0.187 & 2.124 & 72   \\ \hline
8  & 5 & 2.031  & 0.749 & 2.781 & 90   \\ \hline
8  & 6 & 2.078  & 2.999   & 5.078  & 108  \\ \hline
16 & 3 & 3.375 & 0.185 & 3.560 & 216  \\ \hline
16 & 4 & 3.750 & 0.747  & 4.497  & 288  \\ \hline
16 & 5 & 3.938   & 2.997   & 6.935     & 360  \\ \hline
16 & 6 & 4.031  & 11.997  & 16.029 & 432  \\ \hline
\end{tabular}
\end{table}

It can be seen from the configuration combination relationship shown in Table \ref{net_config}. When $BC=16$ and $L=5$, the TBC-Net has 3 million parameters, and the maximum storage requirement is 5 million, which is suitable for most embedded device with CNN accelerators for real-time processing.

\section{Loss Function and Training Method}

The loss function and training method are the core of TBC-Net's ability to learn small target features better. We design a joint loss function for TBC-Net that includes \textbf{t}arget extraction loss, \textbf{b}ackground suppression loss, and \textbf{c}lassification loss to overcome the shortcomings of CNN in learning small target features. This is why we name the network TBC-Net. T, B, and C are the acronyms of the three loss functions. Furthermore, we propose corresponding data synthesis and training methods.

\subsection{Loss Function Design}

\subsubsection{Target Extraction Loss}

The extracted target image $f_{T'}$ should be as close as possible to the ground truth $f_T$. Compared with using $\mathcal{L}_2$ norm to measure the difference between images, $\mathcal{L}_1$ norm and structural similarity ($SSIM$) have better effects \cite{zhao2016loss}\cite{prabhakar2017deepfuse}, but directly using $\mathcal{L}_1$ norm brings halo artifacts, refer to literature \cite{zhao2016loss}, we use the mix loss of $\mathcal{L}_1$ norm and $SSIM$ here, the expression is as follows:

\begin{equation}
\begin{aligned}
&L_{l1}=\frac{1}{N}\sum_{i=1}^{N}||{f_{T'}}_i-{f_T}_i||_{1}\\
&L_S=\frac{1}{N}\sum_{i=1}^{N}1-SSIM({f_{T'}}_i,{f_T}_i)\\
&L_T(f_{T'},f_T)=L_{l1}+L_S\\
\end{aligned}
\end{equation}
where $N$ is the total number of images in the training data, the $SSIM$ index is defined as follows:

\begin{equation}
\begin{aligned}
SSIM(x,y)=\frac{(2\mu_x\mu_y+c_1)(2\sigma_{xy}+c_2)}{(\mu^2_x+\mu^2_y+c_1)(\sigma^2_x+\sigma^2_y+c_2)}
\end{aligned}
\end{equation}
where $\mu_x$, $\sigma_x$ and $\mu_y$, $\sigma_y$ are the pixel mean and standard deviation in the window calculated by sliding a fixed size window on the image $x$ and $y$, respectively. $c_1$, $c_2$ are two variables to stabilize the division with weak denominator.

In experiment, we set $c_1=0.02, c_2=0.06$, and the window size to calculate $\mu_x$, $\sigma_x$ and $\mu_y$, $\sigma_y$ is $11\times 11$.

\subsubsection{Background Suppression Loss}

The TEM obtains an image containing only small targets, and small targets are often sparse in the infrared image. Therefore, we refer to the practice in IPI \cite{gao2013infrared} to use the sparse constraint on the target image to further suppress the background and get a sparse result and the loss can be directly expressed by the $\mathcal{L}_1$ norm of $f_{T'}$ as follows:

\begin{equation}
\begin{aligned}
L_B(f_{T'})=\frac{1}{N}\sum_{i=1}^{N}||{f_{T'}}_i||_{1}=\frac{1}{N}\sum_{i=1}^{N}||TEM({f_D}_i)||_{1}
\end{aligned}
\end{equation}

\subsubsection{Classification Loss}

As mentioned earlier, the semantic constraint module is essentially a network that classifies the target image, so we use the cross-entropy loss expressed as follows:

\begin{equation}
\begin{aligned}
L_C(f_{T'},y_T)&=\frac{1}{N}\sum_{i=1}^{N}CE(SCM({f_{T'}}_i),{y_T}_i)\\
&=\frac{1}{N}\sum_{i=1}^{N}CE(SCM(TEM({f_D}_i)),{y_T}_i)
\end{aligned}
\end{equation}
where $CE$ is the abbreviations of the cross-entropy loss, and $y_T$ denotes the ground truth classification label corresponding to the target image (that is, the number of targets included therein).

\subsubsection{Total loss}
\begin{figure}[!h]
\centering
\includegraphics[width=0.495\textwidth]{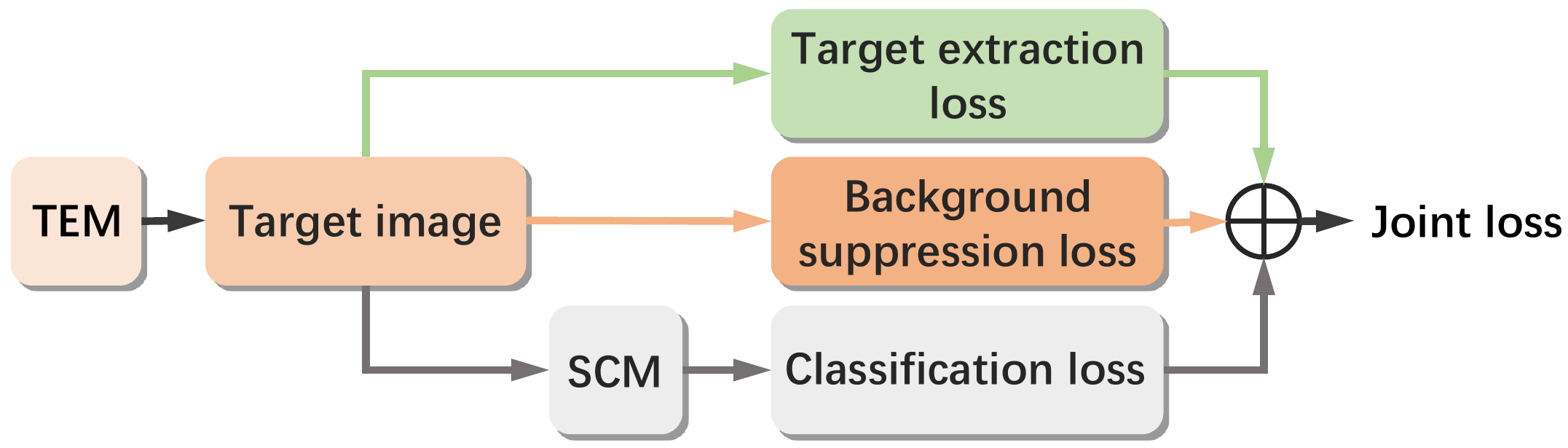}
\caption{The method of calculating joint loss for training TBC-Net.}
\label{cal_loss}
\end{figure}

From the above three loss functions, we can get the joint loss function required to train TBC-Net, as shown below:
\begin{equation}
\begin{aligned}
 L_{TBC}=L_{T} + L_{B} +\lambda L_{C}
\end{aligned}
\end{equation}
where $\lambda$ is the weight of the loss function $L_C$. In the experiment, we take $\lambda=1$. The method to calculate the joint loss during training is illustrated in Figure \ref{cal_loss}.

\subsubsection{Analysis}
Through analysis of the training procedure, we can explain the effectiveness of the joint loss function in small target feature learning. CNN training generally adopts a gradient descent algorithm, that is, the network weights are updated according to the following formula:

\begin{equation}
\begin{aligned}
\omega_{t+1}&=\omega_{t}-\eta\frac{\partial L_{TBC}}{\partial \omega_t}\\
\end{aligned}
\end{equation}
where $\omega_t, \omega_{t+1}, \eta$ denote the network weights in $t$th, $t+1$th training step and the learning rate, respectively.  

With the chain rule, we can decompose the $L_{TBC}$ into four partial derivatives of the output image $f_{T'}$, from which we can see the effect of the joint loss on the TEM output. The decomposition process is as follows:


\begin{equation}
\begin{aligned}
\frac{\partial L_{TBC}}{\partial \omega_{t}}&=\frac{\partial L_{TBC}}{\partial f_{T'}}\frac{\partial f_{T'}}{\partial \omega_{t}} \\
&=(\frac{\partial L_T}{\partial f_{T'}}+\frac{\partial L_B}{\partial f_{T'}}+\frac{\partial L_C}{\partial f_{T'}})\frac{\partial f_{T'}}{\partial \omega_{t}}\\
&=(\frac{\partial L_{l_1}}{\partial f_{T'}}+\frac{\partial L_{S}}{\partial f_{T'}}+\frac{\partial L_B}{\partial f_{T'}}+\frac{\partial L_C}{\partial f_{T'}})\frac{\partial f_{T'}}{\partial \omega_{t}}\\
\end{aligned}
\end{equation}

We combine the four partial derivatives in parentheses into two parts:
\begin{equation}
\begin{aligned}
(\frac{\partial L_{l_1}}{\partial f_{T'}}+\frac{\partial L_{S}}{\partial f_{T'}}+\frac{\partial L_B}{\partial f_{T'}})+\frac{\partial L_C}{\partial f_{T'}}
\end{aligned}
\end{equation}

Because these loss functions are additive relationships, their effects on $f_{T'}$ are independent. We can analyze them separately to explain their effects.

\begin{figure*}[!h]
	\centering
	\includegraphics[width=0.7\textwidth]{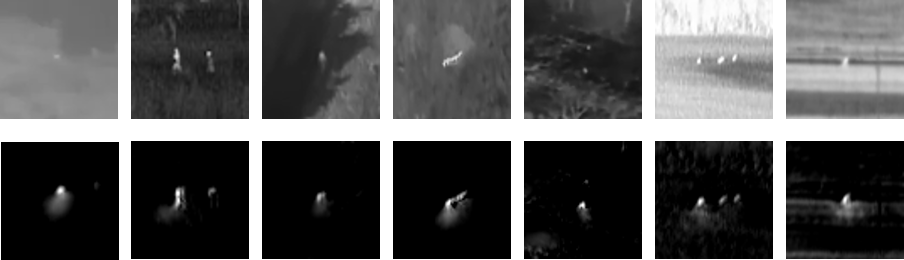}
	\caption{Halo artifacts caused by $L_{l1}$ loss, the first row are original images and the second row are output images $f_{T'}$ of the TEM.}
	\label{halo}
\end{figure*}

\begin{figure}[!h]
	\centering
	\includegraphics[width=0.48\textwidth]{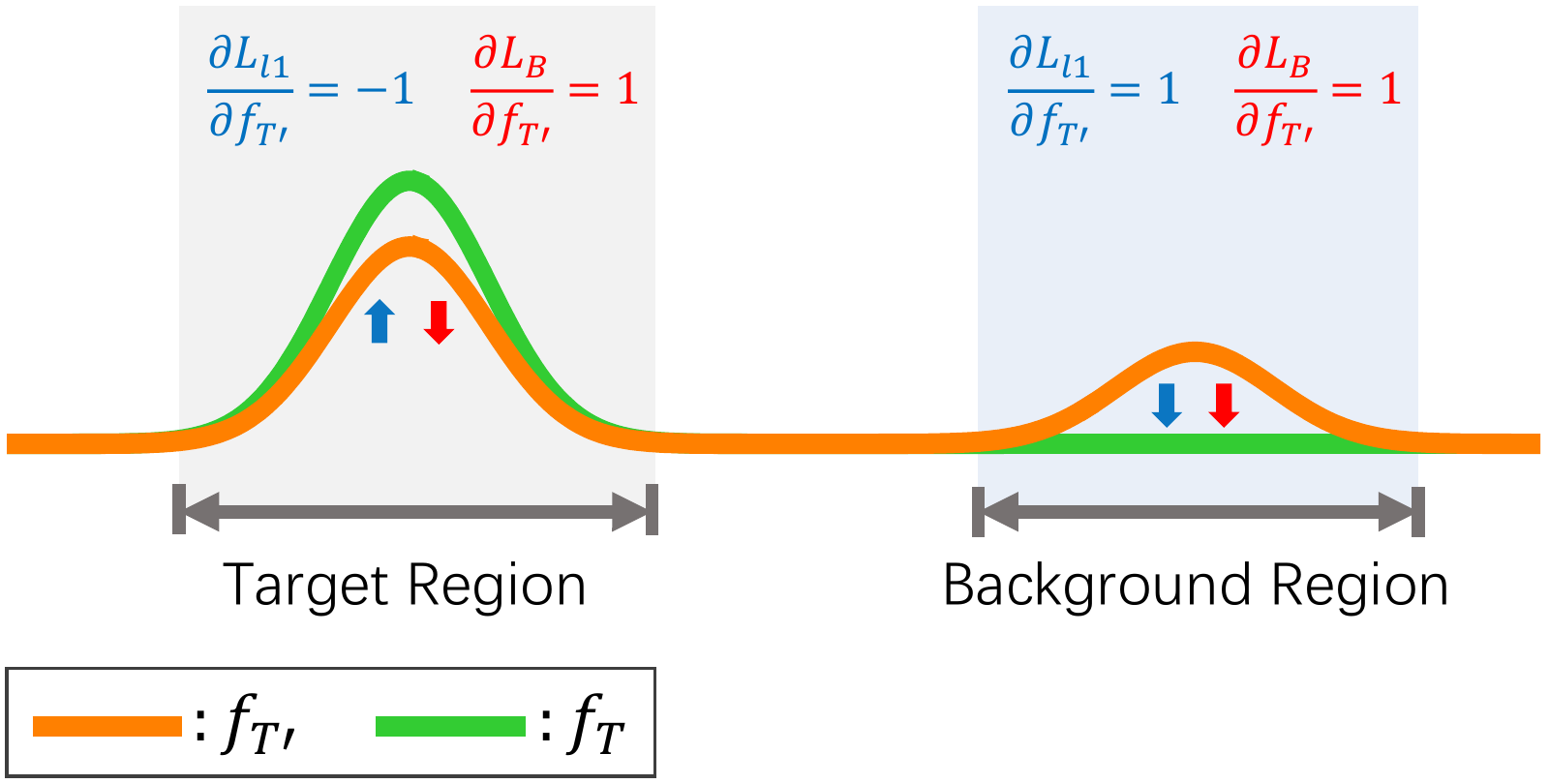}
	\caption{Schematic diagram of qualitative interpretation of the effects of $L_{l1}$ and $L_B$ in the target region and background region (Blue and red arrows represent the effects of $L_{l1}$ and $L_{B}$ gradients on $f_{T'}$, respectively).}
	\label{grad_analysis}
\end{figure}

For the first part, we focus on their joint role in the target region and the background region. In the background region corresponding to the TEM output image $f_{T'}$, if $f_{T'}$ is greater than $f_T$, then $\frac{\partial L_{l_1}}{\partial f_{T'}}$ and $\frac{\partial L_B}{\partial f_{T'}}$ are simultaneously equal to 1, so that the background and fluctuations caused by the update along the $L_C$ gradients direction can be suppressed. In the target region, although the same suppression phenomenon occurs when $f_{T'}$ is greater than $f_T$, when $f_{T'}$ is smaller than $f_{T}$, $\frac{\partial L_{l_1}}{\partial f_{T'}}$ is equal to -1, $\frac{\partial L_B}{\partial f_{T'}}$ is equal to 1, and the gradient of the superposition is 0. Meanwhile, the gradient of $L_S$ still exists, which can make the output $f_{T'}$ of the TEM gradually approach $f_{T}$. The combined effect of $L_{l_1}$ and $L_B$ can shield the gradient update caused by $L_{l_1}$ in the target region to reduce the occurrence of artifacts (as shown in Figure \ref{halo}) without losing the good suppression effect of $L_{l_1}$ on the background region, and there is no need to worry about the target features being smoothed by $L_B$. The qualitative interpretation is shown in Figure \ref{grad_analysis}. In Figure \ref{grad_analysis}, the orange line represents the output grayscale of the TEM, and the green line represents the grayscale of $f_T$. The gray area and light blue area show the combined effect of the gradients of the two loss functions in the target region and the background region, respectively.

Meanwhile, by minimizing the $L_S$, we can make $f_T$ and $f_{T'}$ as close as possible both in the target region and the background region.

For the second part, the classification loss $L_C$ is sensitive to the disappearance of the small target. Once an image with a small targets is processed by the TEM to obtain a blank image without the target, the $L_C$ loss will increase, which can change the $f_{T'}$ and the network can get rid of the current state, which means the SCM generate semantic constraint on the output of the TEM and solve the problem caused by the data imbalance problem.

\subsection{Data Synthesis and Training}

According to the previous design, calculating the loss function requires an original image $f_D$, a target image $f_T$, and a label $y_T$ indicating the number of targets in the original image $f_D$. These three parts constitute a training tuple $(f_D, f_T, y_T)$ for calculating the loss function $L_{TBC}$. We have a large number of background images $f_B$, so the key is to synthesize the image $f_D$ containing several small targets, and give the label $y_T$ according to the number of targets.

\subsubsection{Synthesizing $f_D$}
\begin{figure*}[!h]
	\centering
	\subfloat[target size: 2]{\includegraphics[height=0.26\textwidth]{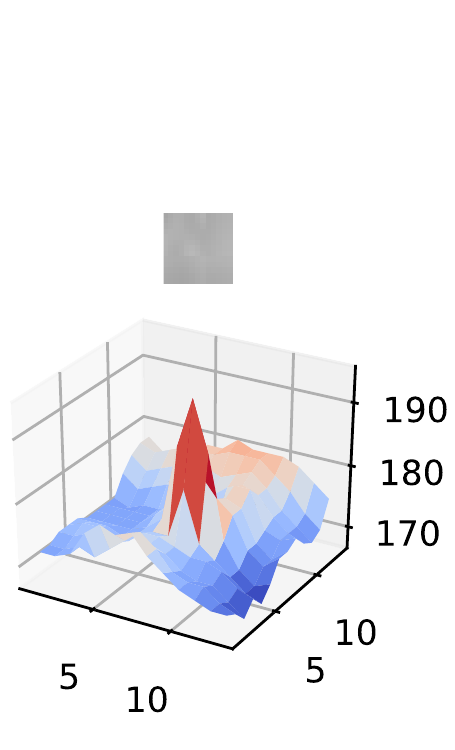}%
	}
	\hfil
	\subfloat[target size: 4]{\includegraphics[height=0.26\textwidth]{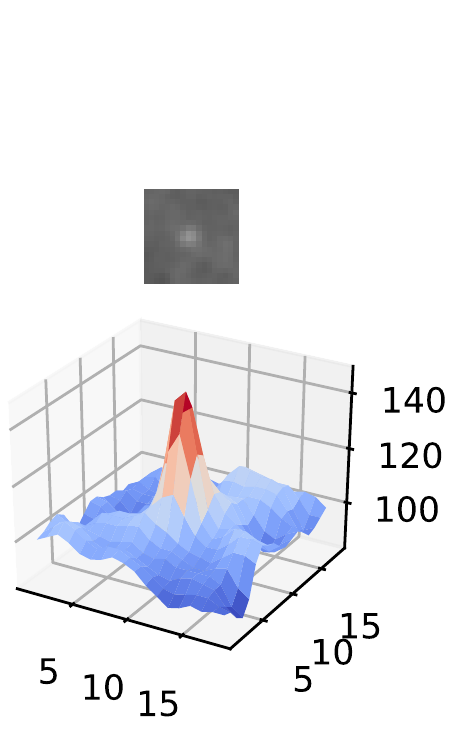}%
	}
	\hfil
	\subfloat[target size: 6]{\includegraphics[height=0.26\textwidth]{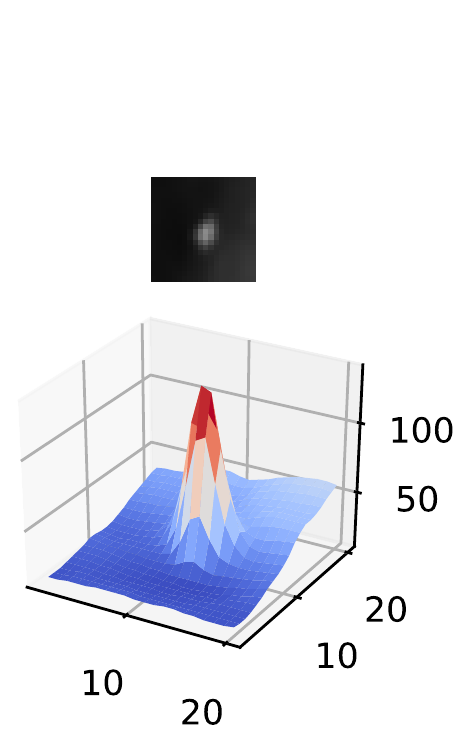}%
}
\hfil
	\subfloat[target size: 8]{\includegraphics[height=0.26\textwidth]{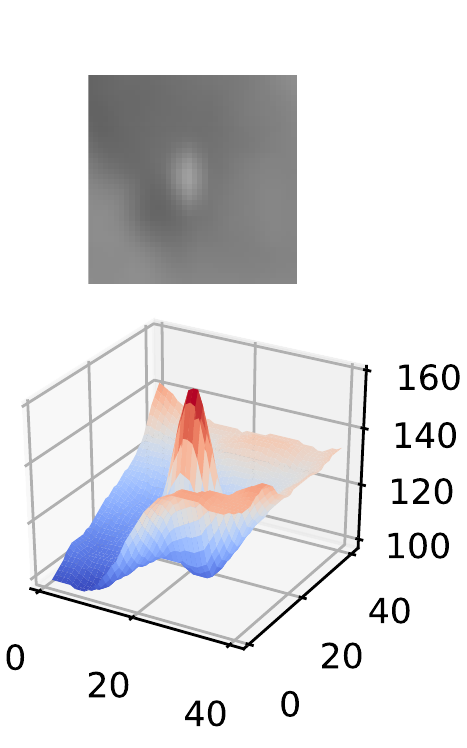}%
}
\hfil
	\subfloat[target size: 10]{\includegraphics[height=0.26\textwidth]{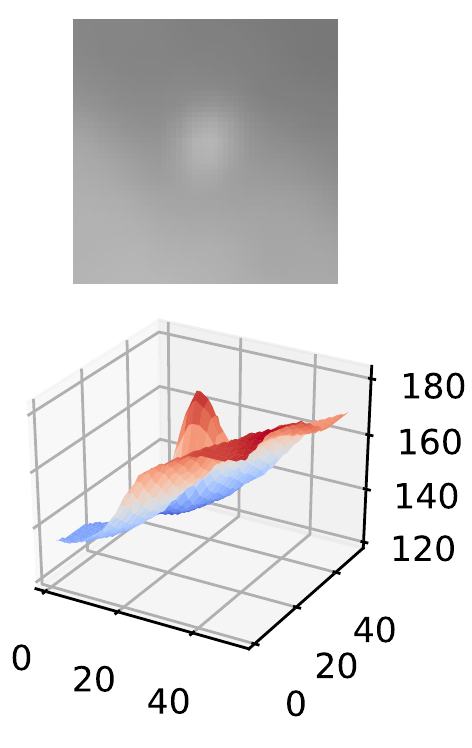}%
}
	\caption{Local structures of small targets in synthesis data.}
	\label{local_structure}
\end{figure*}

Kim \textit{et al.} \cite{kim2014infrared} proposed a method for generating infrared images from object models based on black-body radiation, but we find that when the target size is scaled to within 10 pixels, the shape information of the target itself is weak. Compared to synthesizing data utilizing black-body radiation, it is easier to fuse small target images with the background as in \cite{gao2013infrared}. In more detail, we select one of the target images, randomly adjust the brightness, rescale the target image to a random size, then use the method proposed in \cite{gao2013infrared} to fuse the target into background. The algorithm for fusing a target into the background is shown in Algorithm \ref{fuse_one}, where $\alpha$ is a random factor for adjusting the brightness. Some examples of local structures of small targets that fused into the background are shown in Figure \ref{local_structure}.

 \begin{algorithm}[!h]
	\caption{Algorithm for fusing a target on $f_B$}
	\label{fuse_one}
	\begin{algorithmic}[1]
		\renewcommand{\algorithmicrequire}{\textbf{Input:}}
		\renewcommand{\algorithmicensure}{\textbf{Output:}}
		\REQUIRE Background image $f_B$
		\REQUIRE Target image $t$
		\REQUIRE Fusion location $loc=(x_0,y_0,h_0,w_0)$
		\ENSURE  Fused image $f_{fused}$
		\ENSURE Fused success flag $FusedSuccess$
		\STATE Generate a random brightness adjustment factor $\alpha$ from the uniform distribution of $ [0.75,1]$
		\STATE $f_T=$imresize$(t, (h_0,w_0))$

		
		\STATE $FusedPixels=0$
		\FOR{$x\in (1+x_0,h_0+x_0),y\in (1+y_0,w_0+y_0)$}
		\IF {($\alpha f_T(x-x_0,y-y_0) > f_B(x,y)$)}
		\STATE $f_{fused}(x,y) = \alpha f_T(x-x_0,y-y_0) $
		\STATE $FusedPixels=FusedPixels+1$
		\ELSE 
		\STATE $f_{fused}(x,y)=f_B(x,y)$
		\ENDIF
		\ENDFOR
		\IF {$FusedPixels > 1$}
		\STATE $FusedSuccess=$True
		\ELSE
		\STATE $FusedSuccess=$False
		\ENDIF
		\RETURN $f_{fused}$,$FusedSuccess$ 
	\end{algorithmic} 
\label{fused}
\end{algorithm}

\subsubsection{Label $y_T$}

In the process of synthesizing $f_D$, we mark the corresponding $y_T=n$ according to the number $n$ of small targets successfully fused into the background, and we retain some background images that are not fused with any small target as negative samples and set $y_T=0$, the corresponding training tuple is $(f_B, f_B-f_B, 0)$. When fusing two or more targets into the background, we first generate non-overlapping locations of $n_t$ and fuse $n_t$ different targets into these locations in turn, where $n_t$ is the number of targets we need to add in an image and belongs to $[1, C_{SCM}-1]$. The process is shown in Algorithm \ref{get_tuple}, where $M$ is the total number of images contained in the target image set. Some examples of synthesized data are shown in Table \ref{data_examples}.

In the experimental part, there are 1 to 3 small targets on each synthesis image; that is, $C_{SCM}=4$. Adding up to 3 targets is mainly based on the following two points. On the one hand, when the number of added targets is too small, the classification network is more sensitive to whether there are small targets in the image rather than the number of targets. On the other hand, when too many small targets are added, many of the targets are too close together, which leads to a dense counting problem \cite{boominathan2016crowdnet}\cite{ranjan2018iterative} that is difficult to solve with a classification network. Therefore, according to our attempt, adding 3 to 6 small targets on each image can achieve better results. And the trained classification network can achieve 97.5\% accuracy on predicting the number of targets on $f_T$, which is good enough for guiding the TEM training.

\begin{table*}[!h]
	\renewcommand{\arraystretch}{1.3}
	\centering
	\caption{Examples of background $f_B$, synthesized image $f_D$, target image $f_T$, and label $y_T$.}
	\label{data_examples}
	\begin{tabular}{>{\centering\arraybackslash}m{0.04\textwidth}>{\centering\arraybackslash}m{0.21\textwidth}>{\centering\arraybackslash}m{0.21\textwidth}>{\centering\arraybackslash}m{0.21\textwidth}>{\centering\arraybackslash}m{0.21\textwidth}}
	$f_B$	& \includegraphics*[width=0.17\textwidth]{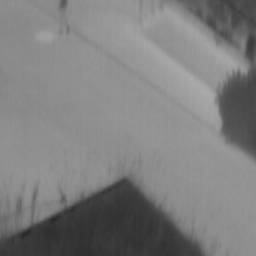}  & \includegraphics*[width=0.17\textwidth]{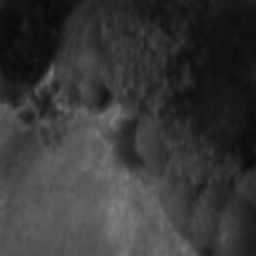}  & \includegraphics*[width=0.17\textwidth]{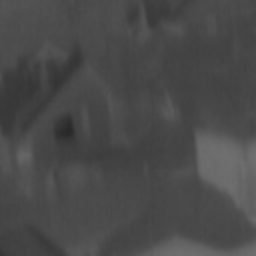} & \includegraphics*[width=0.17\textwidth]{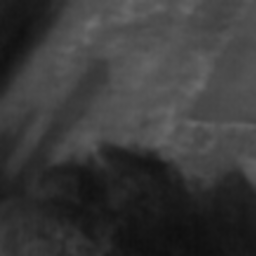} \\
	$f_D$	& \includegraphics*[width=0.17\textwidth]{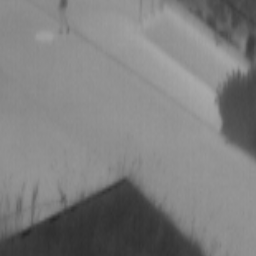} & \includegraphics*[width=0.17\textwidth]{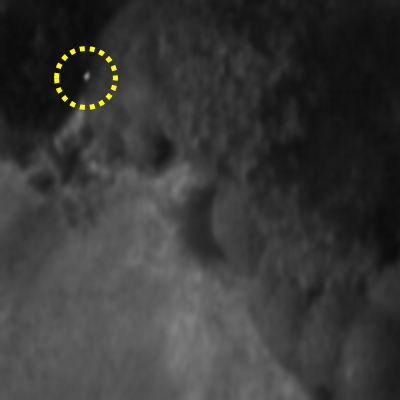} & \includegraphics*[width=0.17\textwidth]{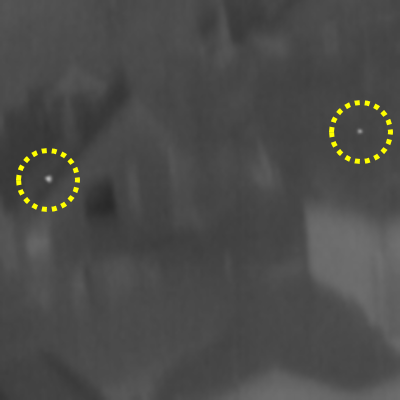} & \includegraphics*[width=0.17\textwidth]{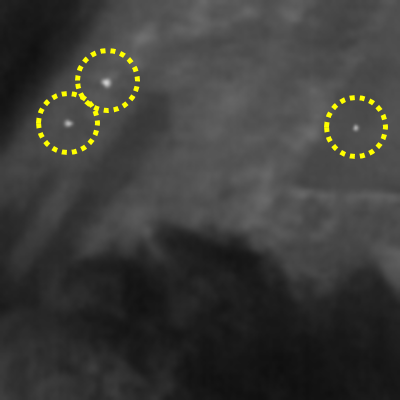} \\
	$f_T$	& \includegraphics*[width=0.17\textwidth]{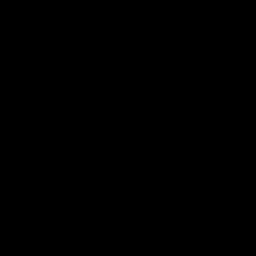} & \includegraphics*[width=0.17\textwidth]{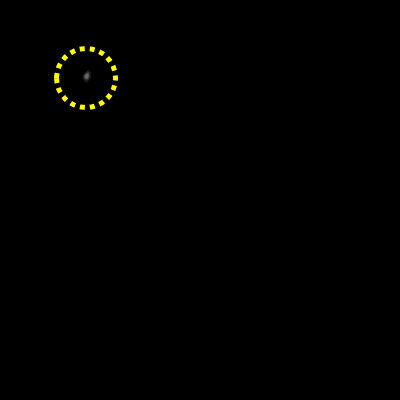} & \includegraphics*[width=0.17\textwidth]{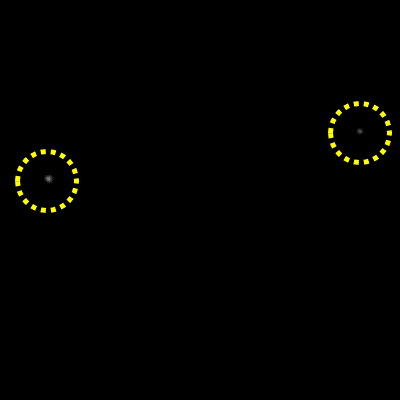} & \includegraphics*[width=0.17\textwidth]{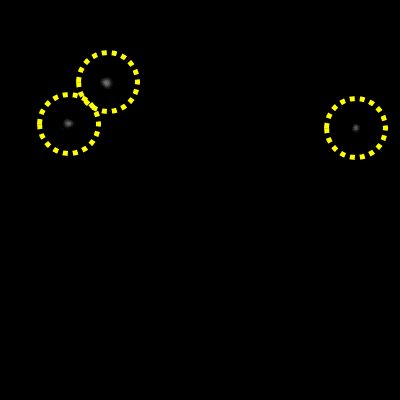} \\
	Label $y_T$	& $y_T=0$ & $y_T=1$ & $y_T=2$ & $y_T=3$
	\end{tabular}
\end{table*}

 \begin{algorithm}[!h]
 \caption{Algorithm for generating a training tuple}
 \label{get_tuple}
 \begin{algorithmic}[1]
 \renewcommand{\algorithmicrequire}{\textbf{Input:}}
 \renewcommand{\algorithmicensure}{\textbf{Output:}}
 \REQUIRE  Background image $f_B$
 \REQUIRE  The number of targets need to fuse $n_t$
 \REQUIRE  Target image set $TS=\{t_1,t_2,...t_M\}$
 \ENSURE  A training tuple $(f_D,f_T,y_T)$
  \STATE $f_D=f_B,y_T=0$
  \STATE Generate $N_t$ non-overlapping locations \\
    $LOC=\{(x_i,y_i,h_i,w_i)\},i=1,2..., n_t$
  \FOR {$loc \in LOC$}
  \STATE Randomly select a target image $t$ from $TS$
  \STATE Use Algorithm \ref{fused} to fuse $t$ on $f_D$ on $loc$
  \STATE Get fused image $f_{fused}$ and flag $FusedSuccess$
  \IF {$FusedSuccess$}
  \STATE $y_T=y_T+1$
  \ENDIF
 \STATE $f_D=f_{fused}$
 \STATE $f_T=f_D-f_B$
  \ENDFOR
 \RETURN $(f_D,f_T,y_T)$ 
 \end{algorithmic} 
 \end{algorithm}

\subsubsection{Training Method}

When training, we take $f_T$ as input and $y_T$ as output ground truth label to train the SCM. When the SCM converges, we freeze its weights, and then use $f_D$ as the input of TBC-Net, and use $L_{TBC}$ to train the TEM. The complete training method of TBC-Net is shown in Algorithm \ref{train_tbc}.

 \begin{algorithm}[!h]
	\caption{Algorithm for training TBC-Net}
	\label{train_tbc}
	\begin{algorithmic}[1]
		\renewcommand{\algorithmicrequire}{\textbf{Input:}}
		\renewcommand{\algorithmicensure}{\textbf{Output:}}
		\REQUIRE  Synthesized training data $\{({f_D}_i, {f_T}_i, {y_T}_i)\}, i=1, 2, ..., N$
		\ENSURE  Trained model TEM
		\STATE Initialize the weights of TEM and SCM.
		\STATE Keep the weights of the TEM not updated
		\STATE Use ${f_T}_i$ as input, ${y_T}_i$ as label and $L_C$ as loss function to train SCM.
		\STATE Keep the weights of the SCM not updated
		\STATE Use training tuples $({f_D}_i, {f_T}_i, {y_T}_i)$ and loss function $L_{TBC}$ to train TEM
		\RETURN Trained model TEM 
	\end{algorithmic} 
\end{algorithm}

\section{Experiments}

The TBC-Net is programmed in Pytorch \cite{paszke2017automatic}. During training phase, the input image size is $256\times256$ with batch size $128$. Adam \cite{kingma2014adam} is used as the optimizer. The initial learning rate is set to 0.005 and decays to one tenth of the previous learning rate at the 80th and 120th epoch for a total of 130 epochs training. The training is done on 4 GTX 1080 GPUs, with an average training time of approximately 65 minutes for every network structure. It should be noted that although the image resolution used in the experiment is $256\times 256$, since the TEM of TBC-Net does not have any constraint on the resolution of the input image, all the layers used can accept input of any resolution. Therefore, TBC-Net can be applied to the detection of infrared small target images of arbitrary resolution. 

Detailed information on the synthesis training data used to train TBC-Net in the experiment is shown in Table \ref{training_data}.

\begin{table}[!h]
	\renewcommand{\arraystretch}{1.3}
	\centering
	\caption{Details of synthesis training data.}
	\label{training_data}
	\begin{tabular}{c|c|c|c|c}
		\hline
		Label $y_T$  & 0    & 1    & 2    & 3    \\ \hline
		Number of images & 2170 & 2402 & 2320 & 2460 \\ \hline
	\end{tabular}
\end{table}

\subsection{Experiment Setup}

\subsubsection{Data Sets}

Six real sequences taken by drones with infrared sensors are used to validate the effectiveness of TBC-Net. These data cover common scenarios and detection issues for UAV field applications such as searching for wildlife and trapped people in the woods. Details about data sets are described in Table \ref{data_info}.

\begin{table*}[!h]
	\renewcommand{\arraystretch}{1.3}
	\centering
	\caption{Details of six infrared small target image data sets}
	\begin{tabular}{ccccc}
		\toprule
		& Frame & Resolution & Target Description                                                            & Background Description \\ \midrule
		Data set 1 & 165   & $256\times 256$    & A person walking in the woods, 3 to 9 pixels in size                            & Woods, buildings       \\
		Data set 2 & 135   & $256\times 256$    & A person walking in the woods, 3 to 9 pixels in size                            & A lot of branches      \\
		Data set 3 & 175   & $256\times 256$    & An animal running towards the woods, 2 to 5 pixels in szie                    & A lot of branches      \\
		Data set 4 & 145   & $256\times 256$    & A person and an animal on the side of the road, 8 pixels and 3 pixels in size & Road, wilderness       \\
		Data set 5 & 60    & $256\times 256$    & An animal running on the edge of the forest, 2 to 3 pixels in size.           & Forest, sunshine       \\
		Data set 6 & 200   & $256\times 256$    & A person walking on a road between the trees, 3 to 8 pixels in size           & Road, woods            \\ \bottomrule
	\end{tabular}
	\label{data_info}
\end{table*}

\subsubsection{Baseline Methods}

We choose three morphological filtering methods Top-Hat, Max-Mean, Max-Median, and five HVS-based methods NWIE, MPCM, HB-MLCM, ILCM, WLCM commonly used in recent years as the baseline methods for experimental comparison. Since \cite{wang2017small} does not disclose the network structure, and \cite{fan2018dim} does not use the network for small infrared target detection, we do not use the \cite{wang2017small}\cite{fan2018dim} methods based on deep learning for comparison. But in the following section, we compare the training method of \cite{liangkui2018using} with our proposed training method. The parameter settings for these methods and TBC-Net are shown in Table \ref{baseline}.

\begin{table*}[!h]
	\renewcommand{\arraystretch}{1.3}
	\centering
	\caption{Parameter settings of different methods}
	\label{baseline}
	\begin{tabular}{cccc}
		\toprule
		No & Method                                       & Abbreviations & Parameter Settings                                                \\ \midrule
		1  & Top-Hat filter \cite{bai2010analysis}       & Top-Hat       & Local window size: $5\times5$                                            \\
		2  & Max-Mean filter  \cite{rivest1996detection}                            & Max-Mean      & Local window size: $15\times15$                                          \\
		3  & Max-Median filter  \cite{rivest1996detection}                          & Max-Median    & Local window size: $15\times15$                                      \\
		4  & High Boost Multiscale Local Contrast Measure \cite{shi2017high}  & HB-MLCM       & External window size: $15\times 15$, target size=${[}3\times 3, 5\times 5, 7\times 7, 9\times 9{]}$ \\
		5  & Improved Local Contrast Measure  \cite{han2014robust}            & ILCM          & Local window size: $8\times 8$ sliding step: $4$                            \\
		6  & Multiscale Patch-based Contrast Measure \cite{wei2016multiscale}      & MPCM          & $L=9$, Local window size: $3\times 3, 5\times 5, 7\times 7$                             \\
		7  & Novel Weighted Image Entropy  \cite{deng2016infrared}               & NWIE          & Local entropy size: $m=7, n=7$; External window size: $7\times 7$            \\
		8  & Weighted new Local Contrast Measure  \cite{liu2018tiny}        & WLCM          & Six structure blocks defined in \cite{liu2018tiny}                      \\
		9  & \textbf{Proposed}                                     & \textbf{Our Method}   & $BC=16, L=5$                                                       \\ \bottomrule
	\end{tabular}
\end{table*}

\subsubsection{Quantative Metrics}

The receiver operating charaterisctic (ROC) curves, signal-to-clutter ratio gain (SCRG), background suppression factor (BSF) are adopted as quantitative evaluation criteria. The ROC curve reflects the trade-off between the detector's detection probability and false alarms and is an important indicator of the quality of a detector. The definitions of detection probability and false-alarm rate are as follows:

\begin{equation}
P_d=\frac{\# real\ targets\ detected}{\# total\ real\ targets}
\end{equation}

\begin{equation}
F_a=\frac{\# false\ pixels\ detected}{\# total\ pixels\ in\ images}
\end{equation}

Depending on the threshold $T$ set by the detector, the detection probability and the false-alarm rate change accordingly, thereby drawing a relationship between the detection probability and the false alarm rate.

The definition of SCRG and BSF are as follows:

\begin{equation}
\begin{aligned}
SCRG&=\frac{SCR_{out}}{SCR_{in}}\\
BSF&=\frac{\sigma_{in}}{\sigma_{out}}\\
\end{aligned}
\end{equation}
where $SCR_{out}$ and $SCR_{in}$ denote signal-to-clutter ratio (SCR) of the output and input images. $SCR=\frac{|M_t-m_b|}{\sigma_b}$, where $M_t$ denotes the pixel value of the target, and $m_b$ and $\sigma_b$ denote the mean and standard deviation of the background area around the target, respectively. In this paper, we use the area of 5 pixels around the target as the background. And $\sigma_{in}$ and $\sigma_{out}$ denote the standard deviation of the background of the image before and after processing, respectively. Since the standard deviation after background suppression may be 0, in order to avoid the denominator being 0, we refer to paper \cite{bai2018derivative}, add an adjustment coefficient $\lambda$ to avoid infinity in SCR and BSF calculations, and set $\lambda$ to 0.01 in this paper.

For convenience, in the experimental part, we normalize the detection results of different algorithms to the $[0,1]$.

\subsection{Ablation Study}

In this section, we compare the network detection performance under different structures and training methods to show the impact of structural changes on the performance of TBC-Net and the effectiveness of our proposed training method.


\subsubsection{Structure Comparison}

In order to simplify the process of exploration, we set the network hyperparameters that need to be explored according to some practical and empirical principles. On the one hand, according to the application scenario of TBC-Net, the number of parameters should be controlled on the order of magnitude similar to the compact neural network \cite{zhang2018shufflenet}\cite{howard2017mobilenets}\cite{iandola2016squeezenet}; that is, the maximum number of parameters should be within 10 million. On the other hand, since the small target size is between 2 and 10 pixels, it is necessary to fuse at least three scales of information to achieve better detection results. Based on the above considerations, we set the $BC$ between 4 and 16, and $L$ between 2 and 5.

By setting different $BC$ and $L$ values, we can get different TBC-Net variants. By testing the ROC curves of different structured TBC-Net variants on real infrared data, we analyze the effects of $BC$ and $L$ on network performance. All of these TBC-Net variants are trained using loss function $L_{TBC}$.

\begin{figure*}[!h]
	\centering
	\subfloat[]{\includegraphics[width=0.243\textwidth]{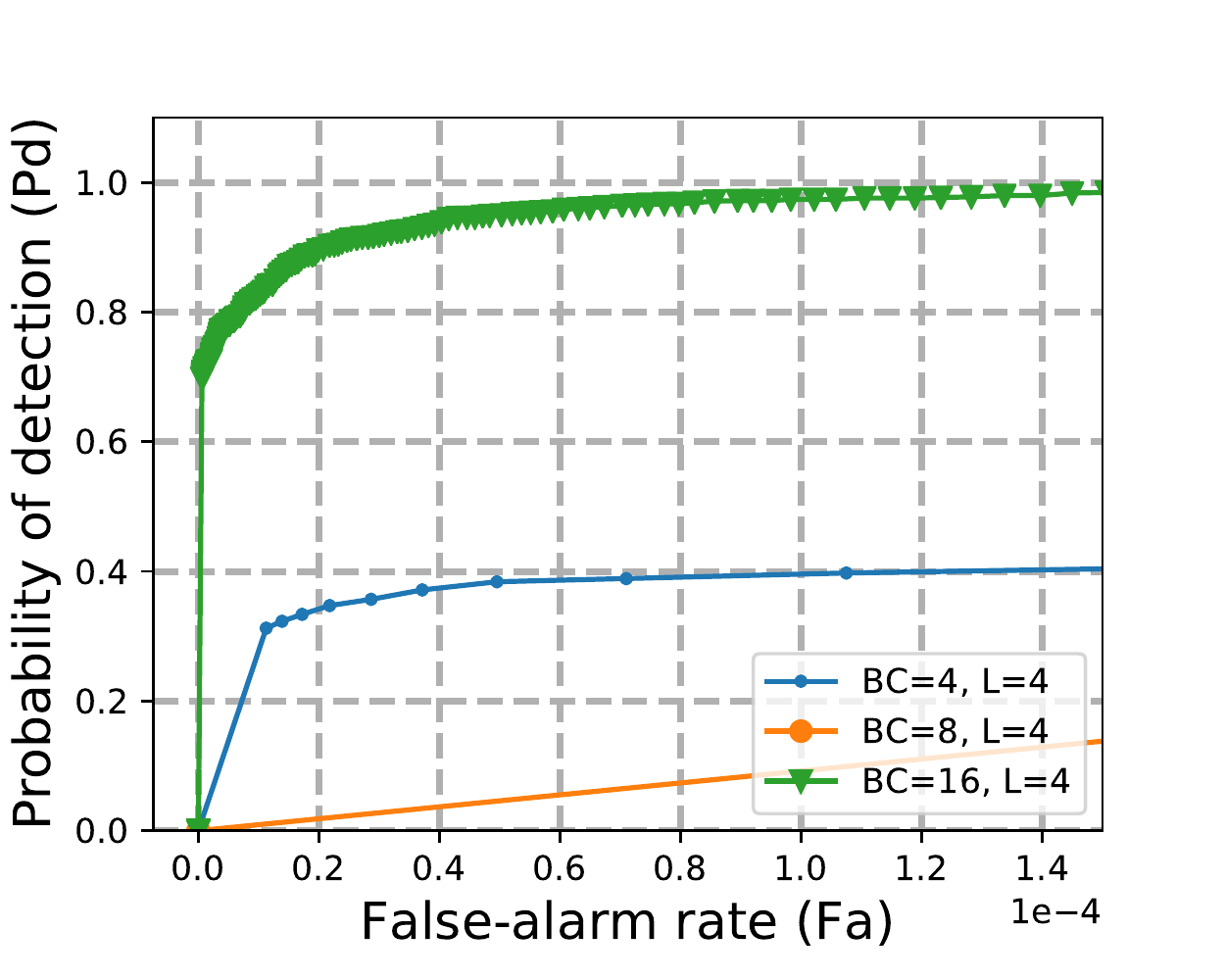}%
		\includegraphics[width=0.243\textwidth]{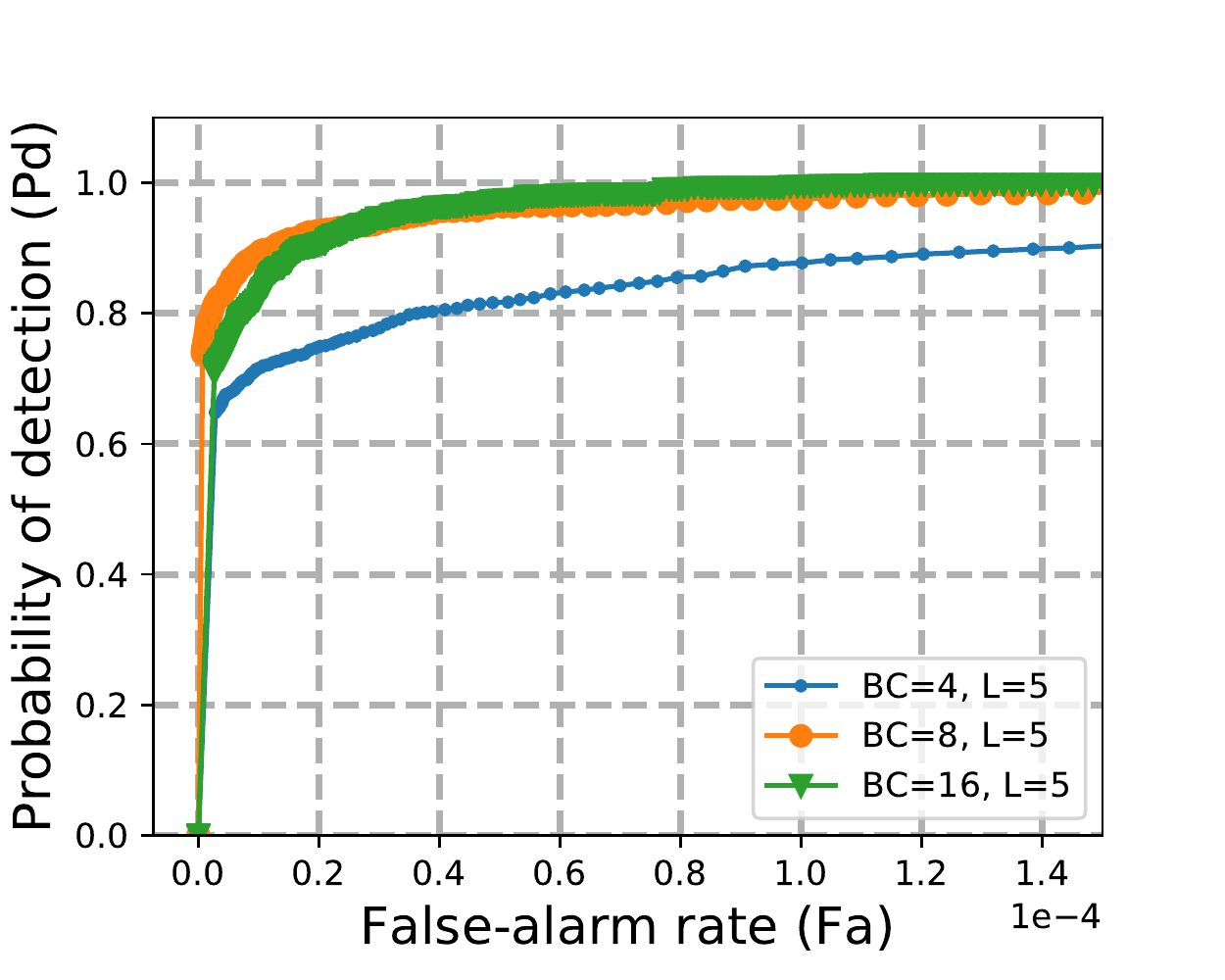}%
		\label{bc_test}
		}
	\hfil
	\subfloat[]{\includegraphics[width=0.243\textwidth]{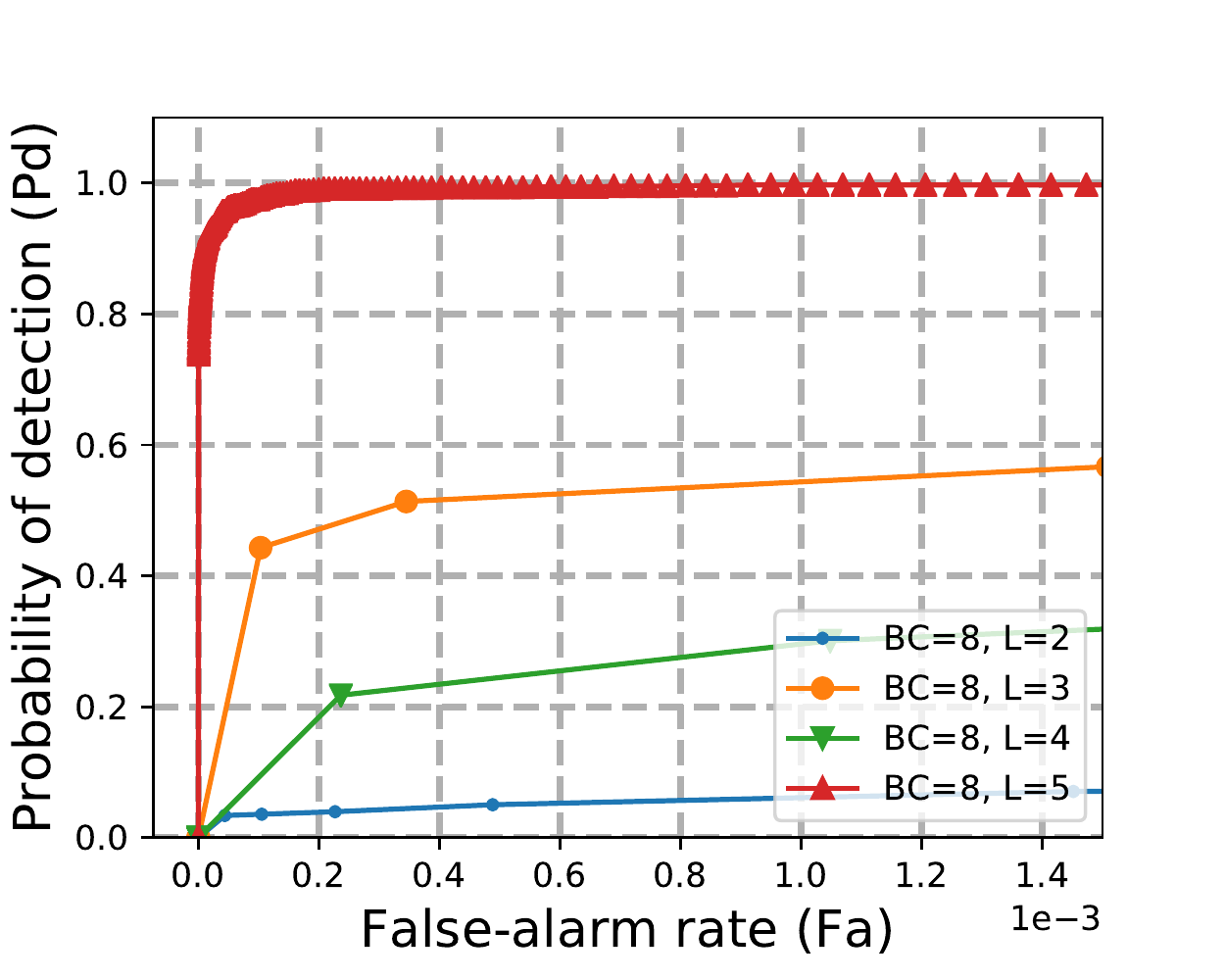}%
		\includegraphics[width=0.243\textwidth]{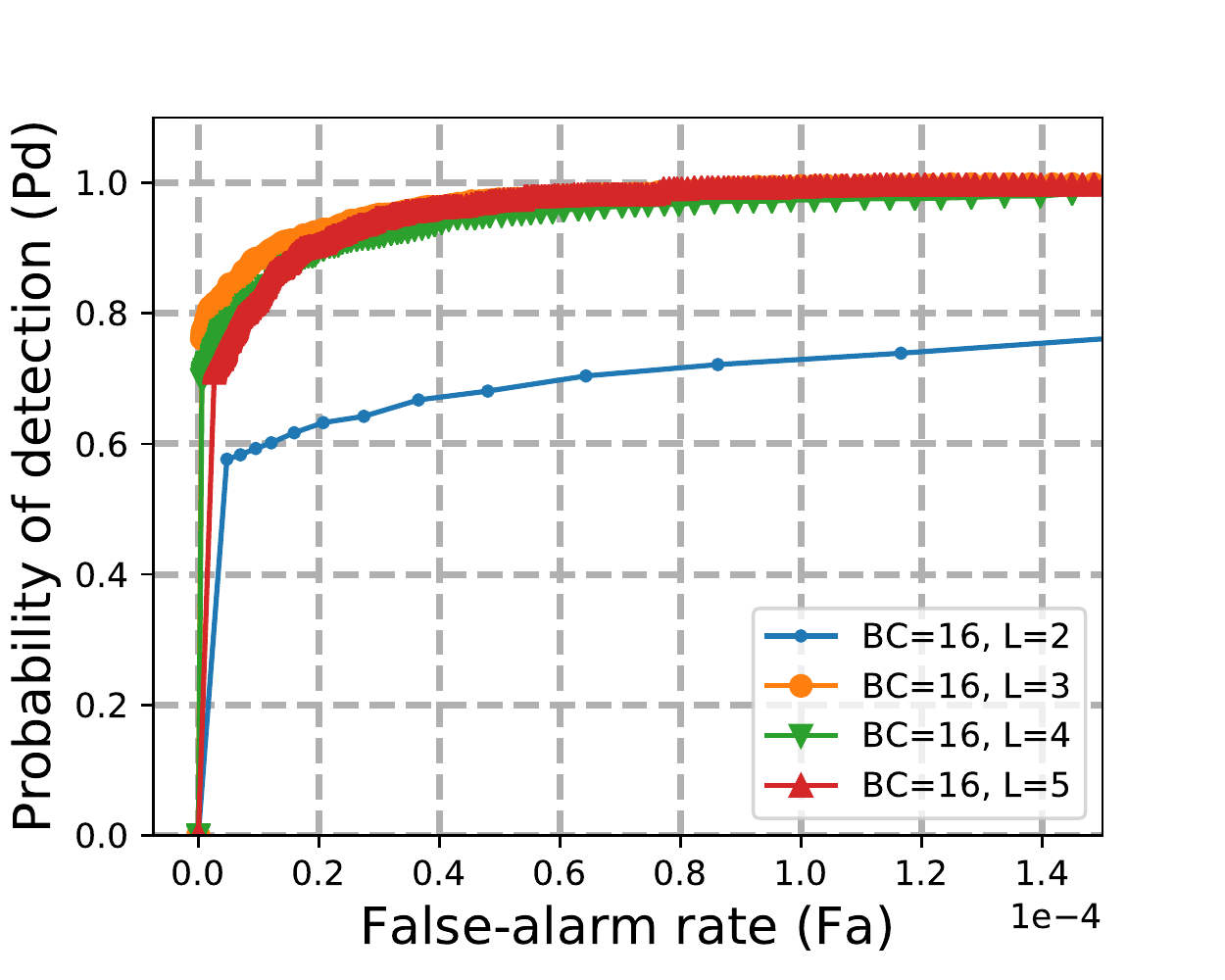}
		\label{l_test}
		}
	\caption{The receiver operating characteristic (ROC) curves of different TBC-Net variants (a) Keep $L$ and change $BC$ (b) Keep $BC$ and change $L$}
\end{figure*}

Keep $L$ unchanged, the ROC curve obtained by changing the value of $BC$ is shown in Figure \ref{bc_test}. It can be seen that when $L=4$ and $5$, the detection performance of TBC-Net improves while increasing $BC$, but when $L=4$, the network is more sensitive to the change of $BC$. When $L=5$, the difference between $BC=8$ and $BC=16$ is not obvious.

Keep $BC$ unchanged, change the value of $L$ to get the ROC curve as shown in Figure \ref{l_test}. Similar to Figure \ref{bc_test}, when $BC=8$, the network is more sensitive to $L$ changes, and when $BC=16$, except for $L=2$, the network performance when $L\ge 3$ is very close.

Structure analysis shows that when one of $BC$ and $L$ is small, the network performance is very sensitive to changes in the other. When $BC = 16$ and $L = 5$, the network performance is good enough.

\subsubsection{Training Method Comparison}

Six loss functions are designed to compare the detection performance of the network trained under different loss function combinations. The six loss functions are $L_T$, $L_T+L_B$, $L_T+L_C$, $L_C$, $L_C+L_B$ and $L_{TBC}$, where $L_{T}$ is widely used in many natural image segmentation and denosing tasks which can be compared as a baseline training method. Other combinations are used to prove the effectiveness of our proposed method. We set $ BC=16, L=5$ to reduce the exploration space.

\begin{figure*}[!h]
	\centering
	\subfloat[Original image]{\includegraphics[width=0.14\textwidth]{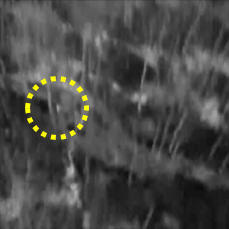}%
		\label{example}}
	\hfil
	\subfloat[]{\includegraphics[width=0.14\textwidth]{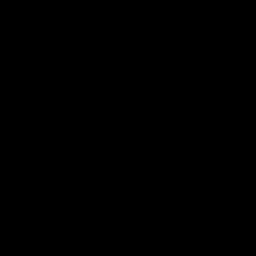}%
		\label{res_T}}
	\hfil
	\subfloat[]{\includegraphics[width=0.14\textwidth]{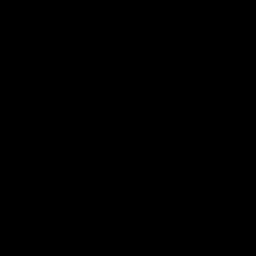}%
		\label{res_TB}}
	\hfil
	\subfloat[]{\includegraphics[width=0.14\textwidth]{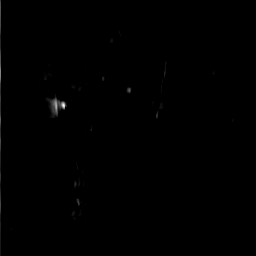}%
		\label{res_TC}}
	\hfil
	\subfloat[]{\includegraphics[width=0.14\textwidth]{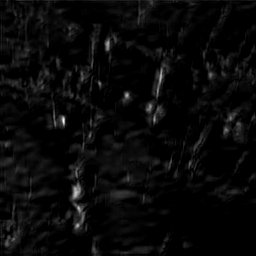}%
		\label{res_C}}
	\hfil
	\subfloat[]{\includegraphics[width=0.14\textwidth]{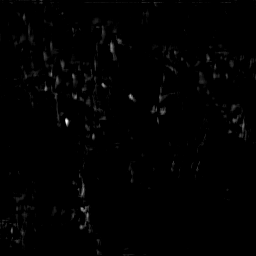}%
		\label{res_CB}}
	\hfil
	\subfloat[]{\includegraphics[width=0.14\textwidth]{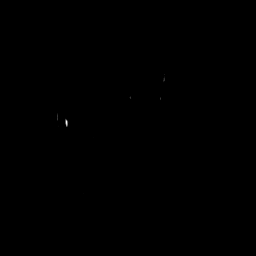}%
		\label{res_CBT}}
	\caption{Examples of output images obtained by networks trained under different loss functions (b) Trained under $L_T$ (c) Trained under $L_T$ and $L_B$ joint loss (d) Trained under $L_T$ and $L_C$ joint loss (e) Trained under $L_C$ (f) Trained under $L_C$ and $L_B$ joint loss (g) Trained under $L_{TBC}$}
	\label{pic_comparison}
\end{figure*}

We take an image in data set 2 as an example to illustrate the difference in the target image obtained under different loss function training, as shown in Figure \ref{pic_comparison}. It can be seen that on the data set 2 with a large amount of clutters and noise in the background, \ref{res_T} and \ref{res_TB} have missed detection. Although networks trained under $L_T$ and $L_T+L_B$ have a strong suppression of the background, they fail to enhance the target. The network trained under $L_T+L_C$ can learn the features of the small target, but there are some halo artifacts around the small target. Networks trained under $L_C$ and $L_C+L_B$ can enhance the target mixed in clutters, but there is a lot of halo artifacts in \ref{res_C}, \ref{res_CB} is better than \ref{res_C} but there is still a lot of noise in the background. The network output \ref{res_CBT} under $L_{TBC}$ training not only suppresses clutters and noise well, but also enhances the target. Besides, we can see in \ref{res_C} that even if only $L_C$ is used, the network can still learn certain target features, which shows that the SCM does have a guiding effect on the TEM during the training phase.

\begin{figure}[!h]
	\centering
	\includegraphics[width=0.33\textwidth]{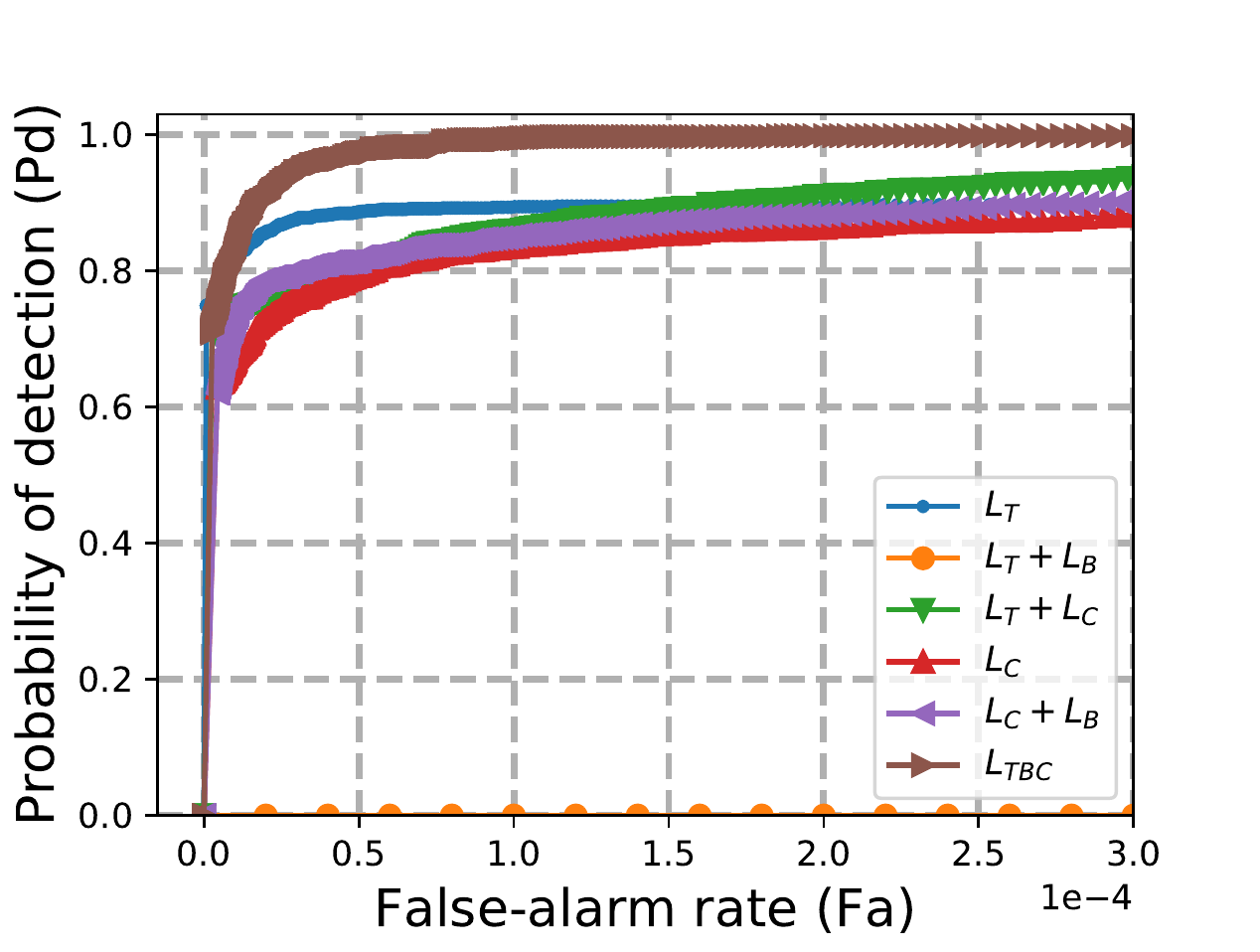}
	\caption{ROC curve of TBC-Net ($BC=16, L=5$) trained under different loss functions tested on all data sets.}
	\label{loss_comparison}
\end{figure}

We use all the images of 1 to 6 data sets to test the network under different loss function training, and plot the ROC curves as shown in Figure \ref{loss_comparison}. It can be seen that the detection performance under $L_{TBC}$ training is the best, while the networks obtained under the training of other loss functions have serious missed detection.

\begin{figure}[!h]
	\centering
	\includegraphics[width=0.34\textwidth]{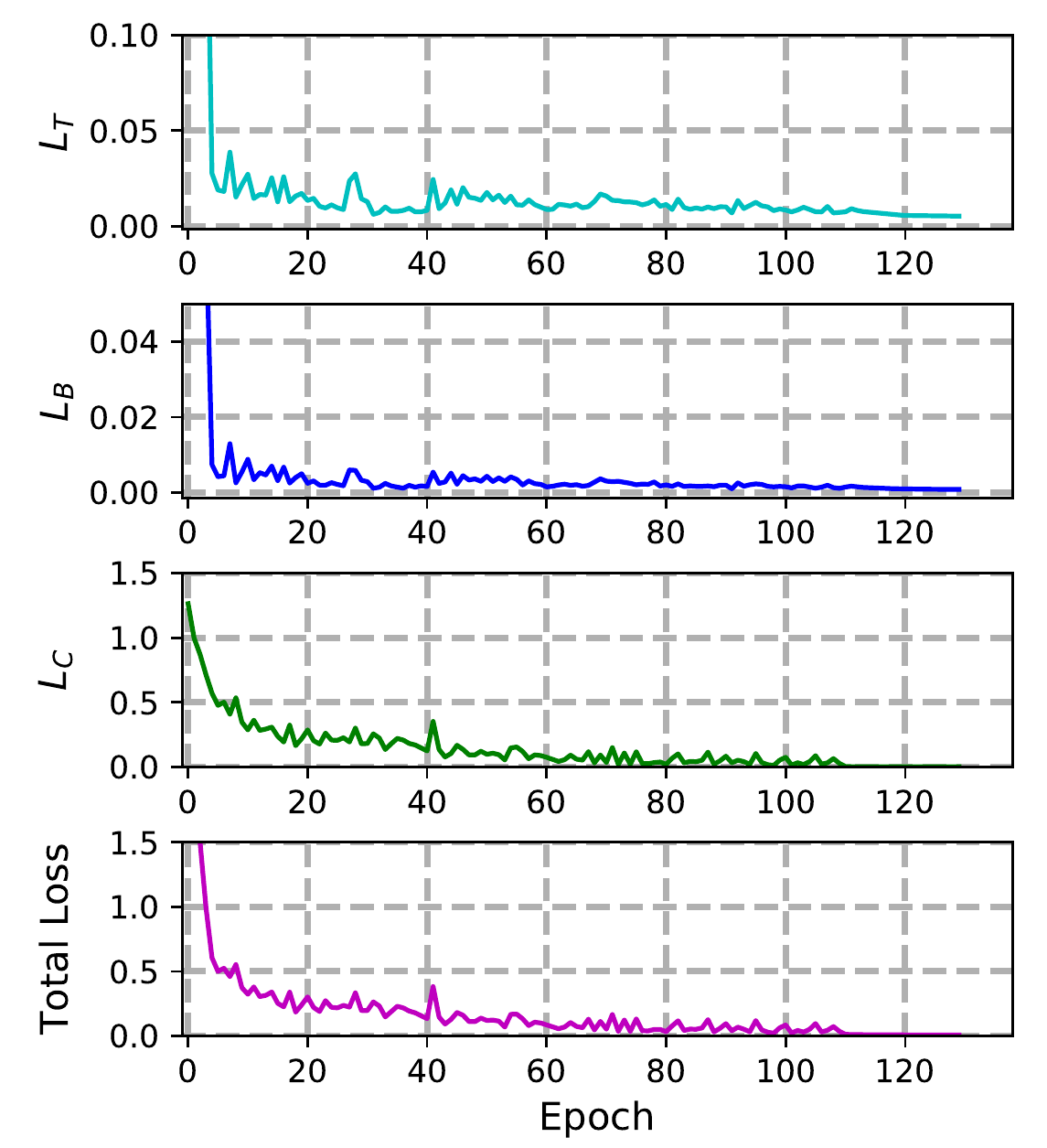}
	\caption{Loss curve of training TBC-Net ($BC=16, L=5$) with $L_{TBC}$}
	\label{loss_curve_TBC}
\end{figure}

\begin{figure*}[!h]
	\centering
	\subfloat[Original image]{\includegraphics[width=0.14\textwidth]{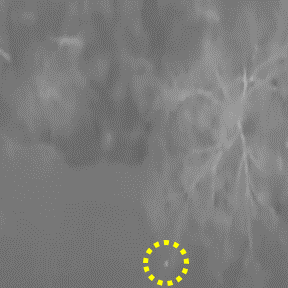}%
		\label{origin}}
	\hfil
	\subfloat[Epoch=5]{\includegraphics[width=0.14\textwidth]{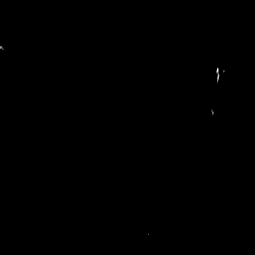}%
		\label{epoch_5}}
	\hfil
	\subfloat[Epoch=10]{\includegraphics[width=0.14\textwidth]{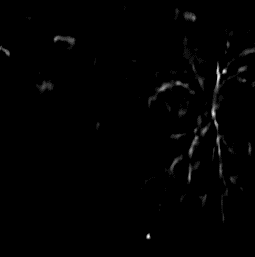}%
		\label{epoch_10}}
	\hfil
	\subfloat[Epoch=50]{\includegraphics[width=0.14\textwidth]{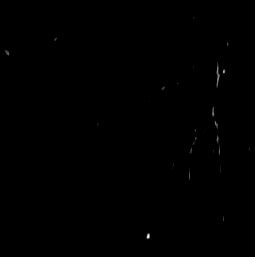}%
		\label{epoch_50}}
	\hfil
	\subfloat[Epoch=100]{\includegraphics[width=0.14\textwidth]{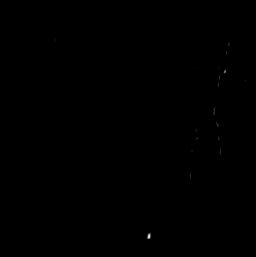}%
		\label{epoch_100}}
	\hfil
	\subfloat[Epoch=130]{\includegraphics[width=0.14\textwidth]{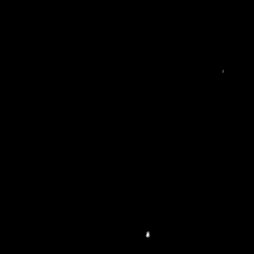}%
		\label{epoch_130}}
	\caption{Output examples of TBC-Net at different epochs under $L_{TBC}$ training}
	\label{epoch_comparison}
\end{figure*}

In addition, we give a training loss curve of TBC-Net trained under $L_{TBC}$, as shown in Figure \ref{loss_curve_TBC}. The total loss in the figure is $L_{TBC}$, and $L_T$, $L_B$, and $L_C$ are the three parts of $L_{TBC}$. In Figure \ref{loss_curve_TBC}, the loss functions do not decay to 0 as quickly as in Figure \ref{loss_example}, but instead approach 0 during continuous adjustment. In fact, although the $L_{TBC}$ does not always approach 0, under the combined effect of several losses, the target extraction and detection is better than the network obtained by overfitting the background. We use the network parameters at different epochs trained under $L_{TBC}$ to test the real data, and use an image in data set 3 as an example to show the change in the small target detection effect of the network during training. The results are shown in Figure \ref{epoch_comparison}. It can be seen that at the beginning, TBC-Net is unable to distinguish the position of the small target, but with the progress of training, TBC-Net can learn the features of the small target more and more clearly, and it also better suppresses the cluttered background.

The comparative analysis of the loss function shows that the $L_{TBC}$ helps to solve the problem of difficult to learn small target features caused by the imbalance between the target and background data. Meanwhile, $L_{TBC}$ has a better effect on suppressing background and artifacts.

\subsection{Qualitative Evaluation with Baseline Methods}

We select a image from each of the six data sets for qualitative evaluation. The original image and the 3D gray-scale distribution of images processed by different algorithms are shown in Figure \ref{seq01} to \ref{seq06}, respectively. The red rectangle is the enlargement of the target region.

For data set 1, both MPCM and TBC-Net have significant enhancements to the target in the forest, but the background suppression of MPCM is worse than TBC-Net, and other methods respond more strongly to the building than to the target. For data set 2, ILCM, NWIE, MPCM, HB-MLCM, and TBC-Net all enhance the target in the woods, but TBC-Net shows greater suppression of the background. For data set 3, in addition to MPCM, other methods can enhance the target running to the woods, and TBC-Net can still suppress the interference of the woods well. For data set 4, there is interference at the edge of the road, and although all methods can enhance both targets, TBC-Net is less sensitive to road edges than other methods. For data set 5, most methods are interfered by the noise of the forest edge and light, which affects the detection of the target, and TBC-Net can better suppress the background noise and enhance the target. For data set 6, all methods have enhanced the target, but NWIE is disturbed by the woods, MPCM produces a strong response on both sides of the road, and TBC-Net can better suppress the interference caused by trees and roads.

\begin{figure*}[!h]
	\centering
	\hfil
	\subfloat[]{\includegraphics[width=0.86\textwidth]{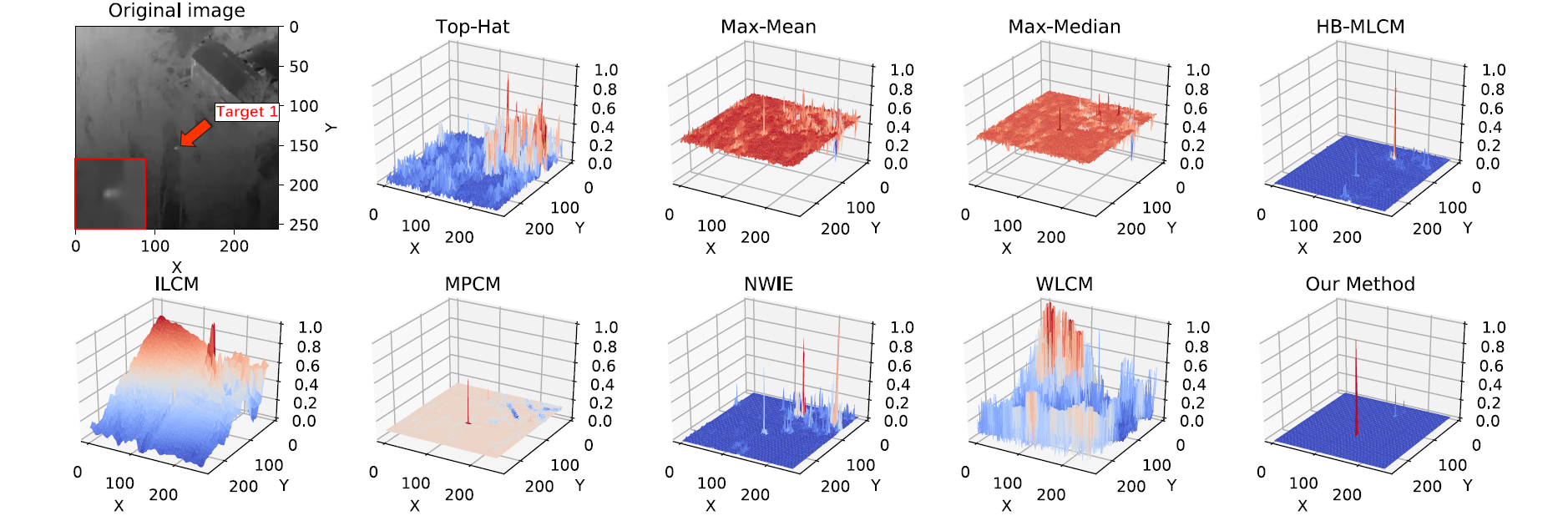}%
		\label{seq01}}\\
	\hfil
	\subfloat[]{\includegraphics[width=0.86\textwidth]{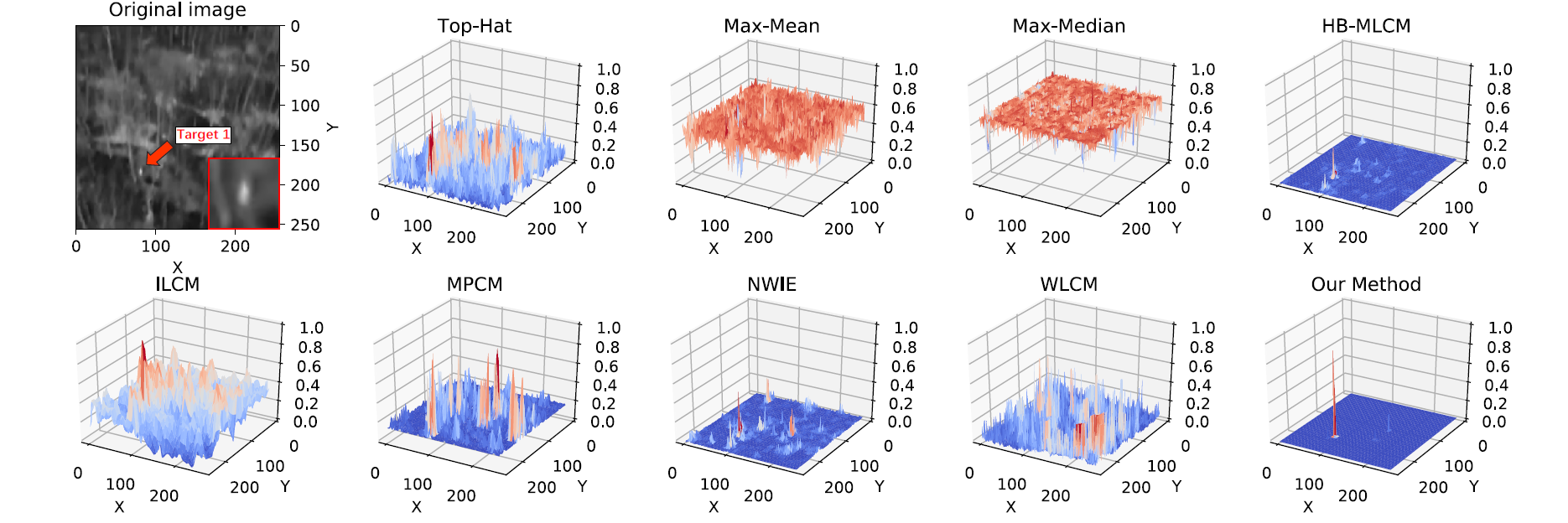}%
		\label{seq02}}\\
	\hfil
	\subfloat[]{\includegraphics[width=0.86\textwidth]{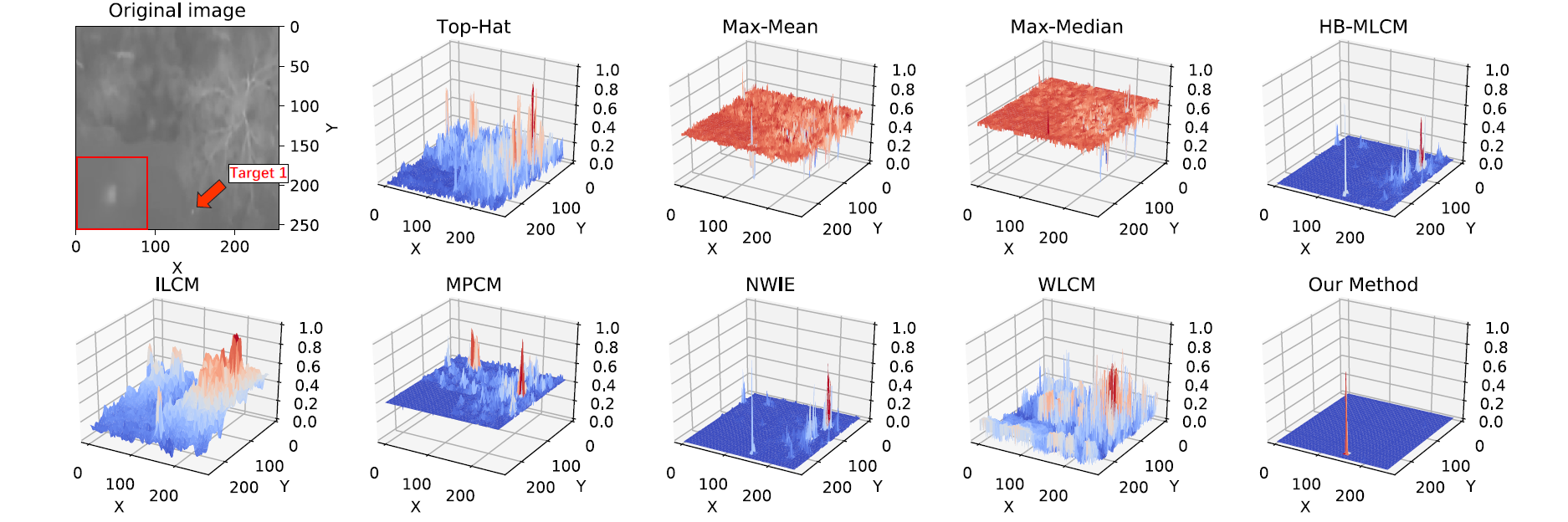}%
		\label{seq03}}\\
	\caption{Original images of data set 1-6 and 3D gray distributions output by different algorithms. (a) Data set 1 (b) Data set 2 (c) Data set 3 (d) Data set 4 (e) Data set 5 (f) Data set 6}
\end{figure*}

\begin{figure*}[!h]
	\ContinuedFloat
	\centering
	\hfil
	\subfloat[]{\includegraphics[width=0.86\textwidth]{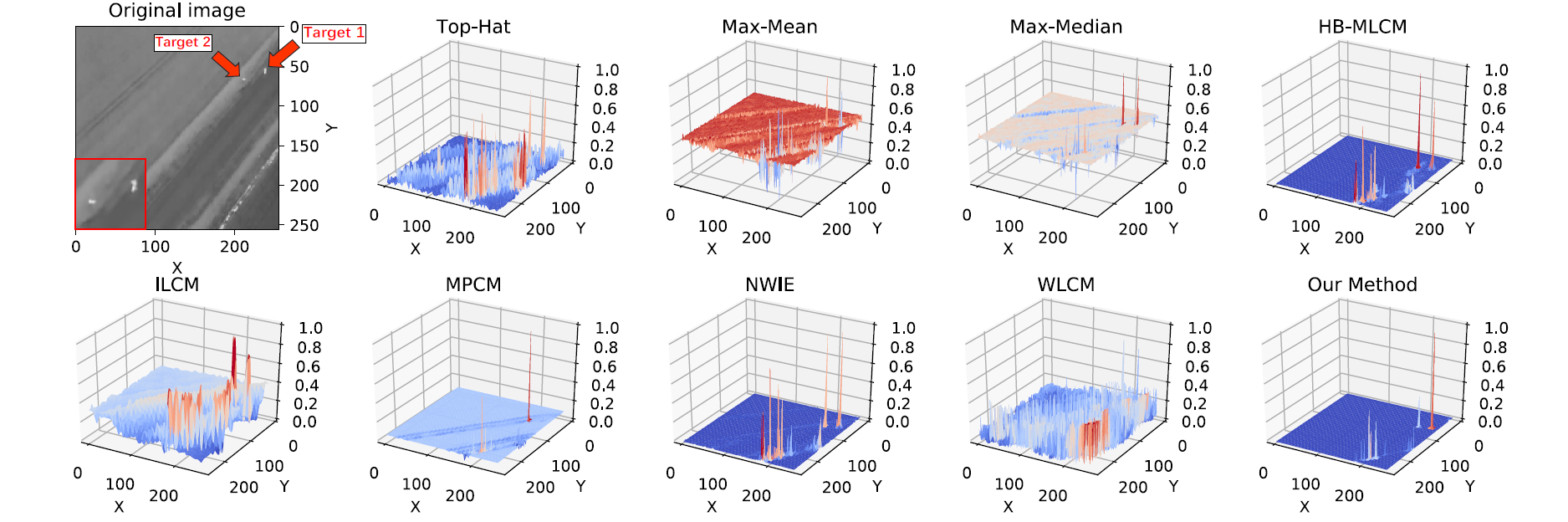}%
		\label{seq04}}\\
	\hfil
	\subfloat[]{\includegraphics[width=0.86\textwidth]{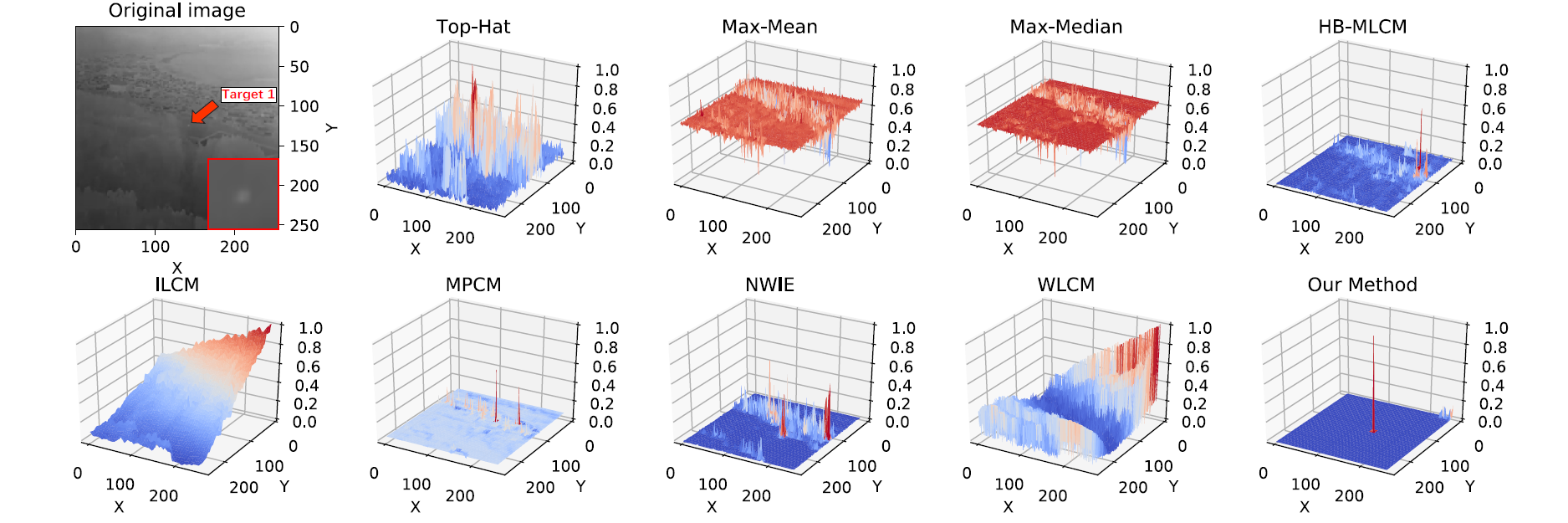}%
		\label{seq05}}\\
	\hfil
	\subfloat[]{\includegraphics[width=0.86\textwidth]{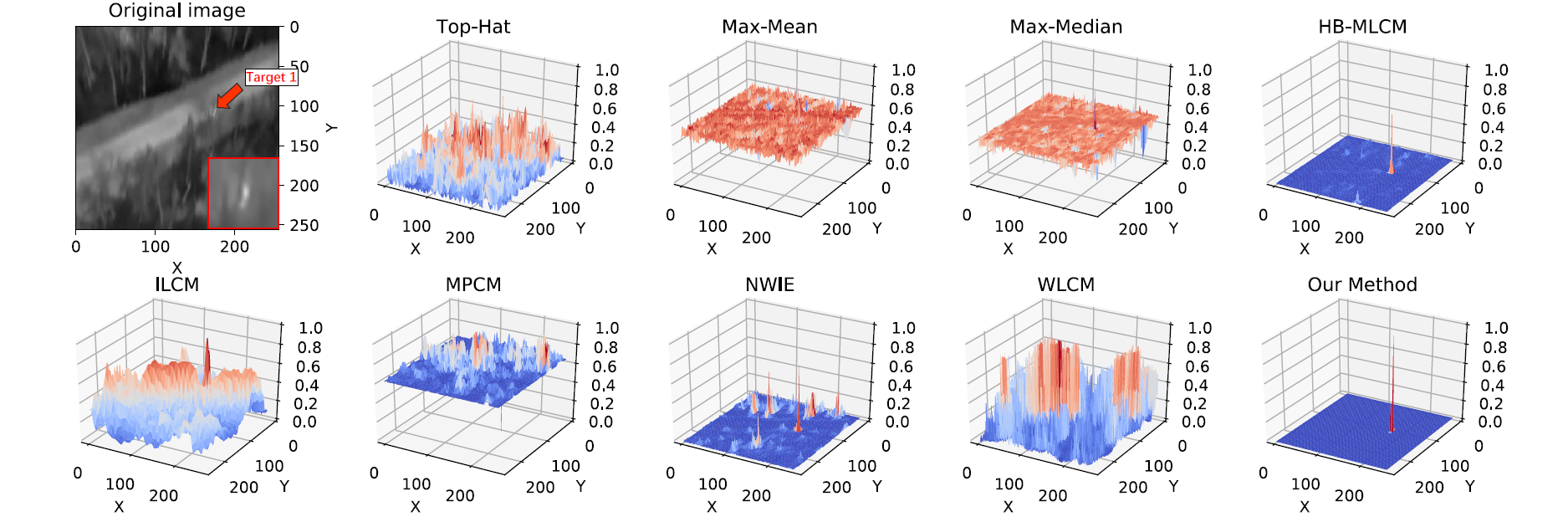}%
		\label{seq06}}
	\caption{Original images of data set 1-6 and 3D gray distributions output by different algorithms. (a) Data set 1 (b) Data set 2 (c) Data set 3 (d) Data set 4 (e) Data set 5 (f) Data set 6}
\end{figure*}

\subsection{Quantative Evaluation with Baseline Methods}

\begin{figure*}[!h]
	\centering
	\offinterlineskip
	\subfloat[]{\includegraphics[width=0.33\textwidth]{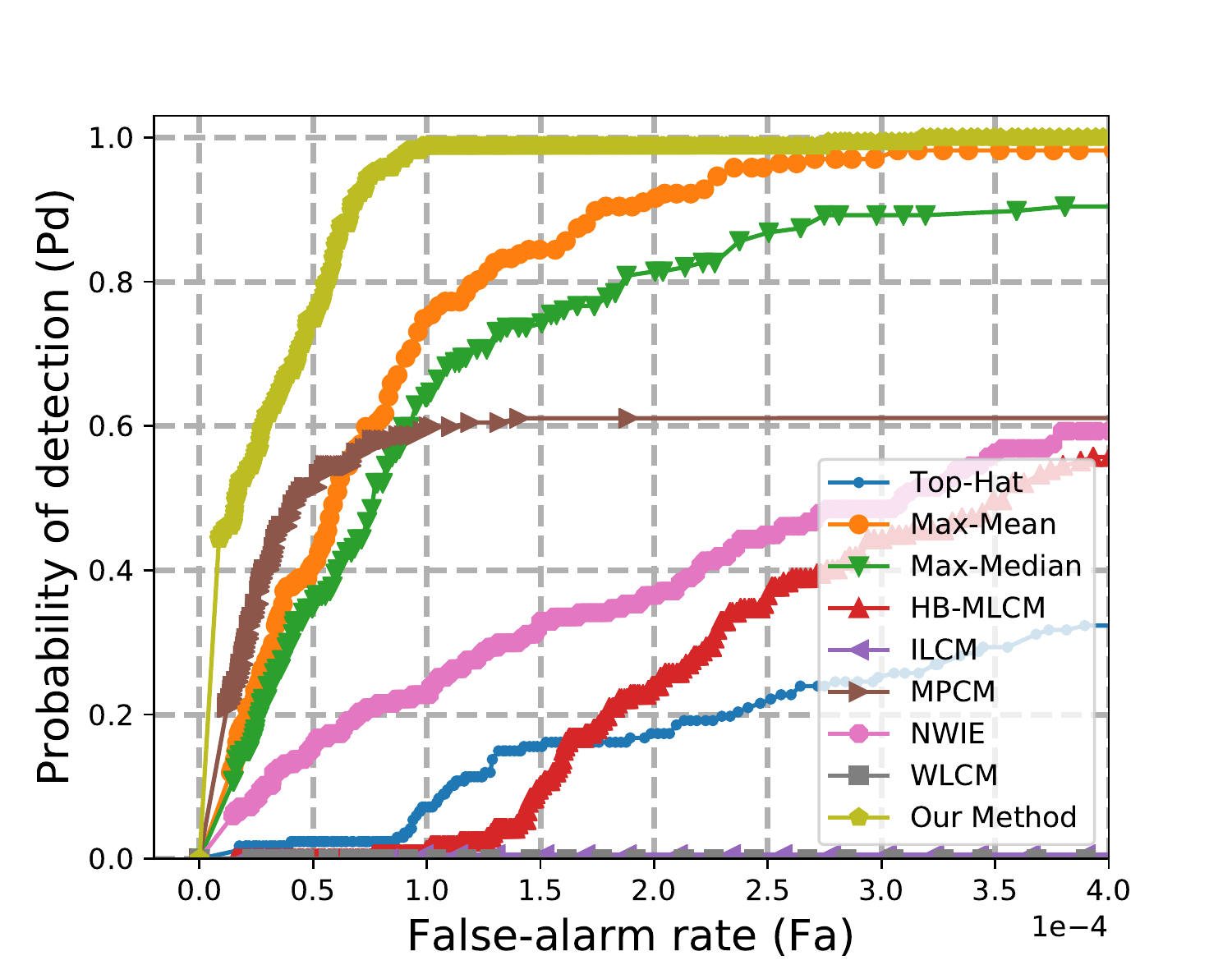}%
		\label{seq_roc01}}
	\hfil
	\subfloat[]{\includegraphics[width=0.33\textwidth]{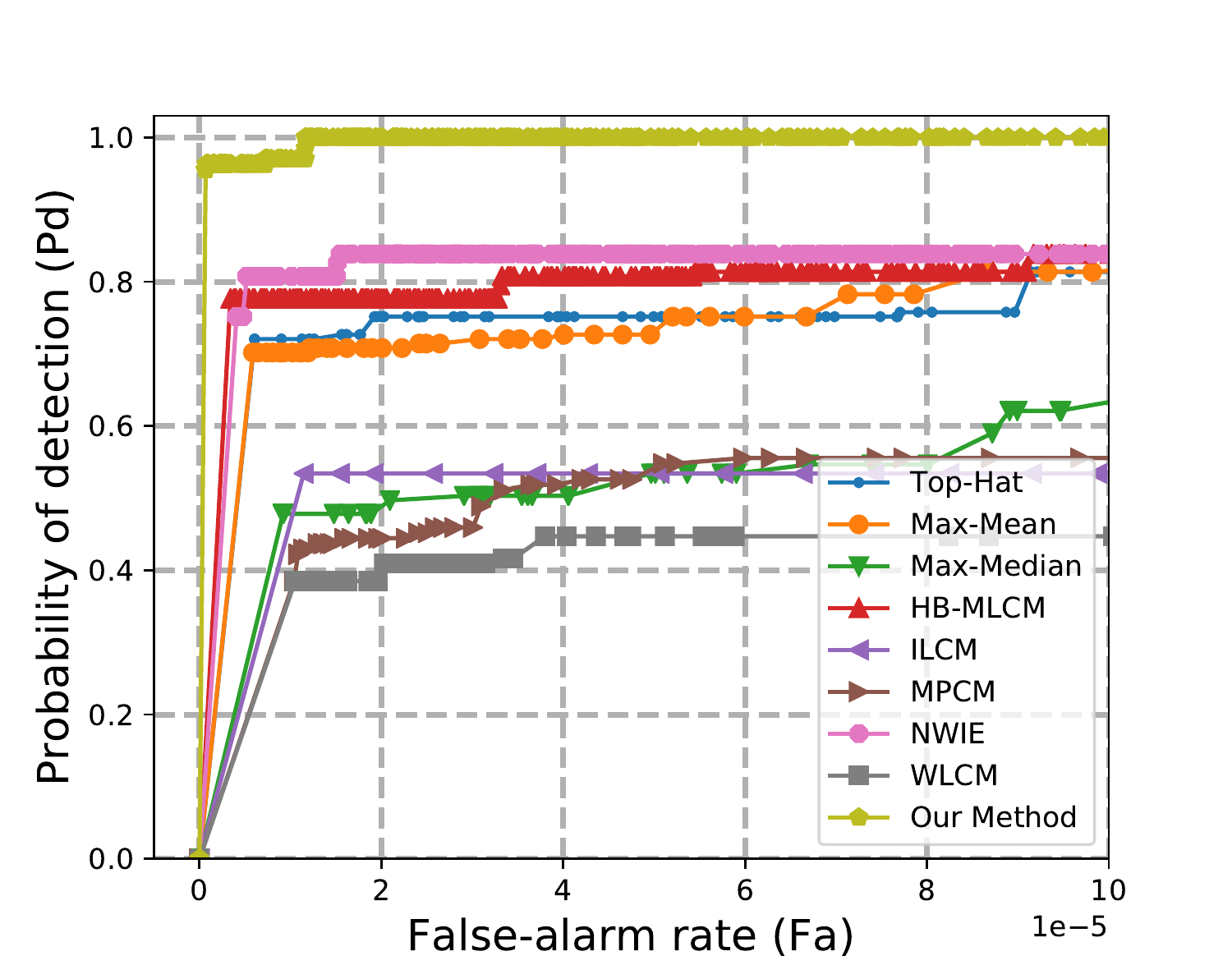}%
		\label{seq_roc02}} 
	\hfil
	\subfloat[]{\includegraphics[width=0.33\textwidth]{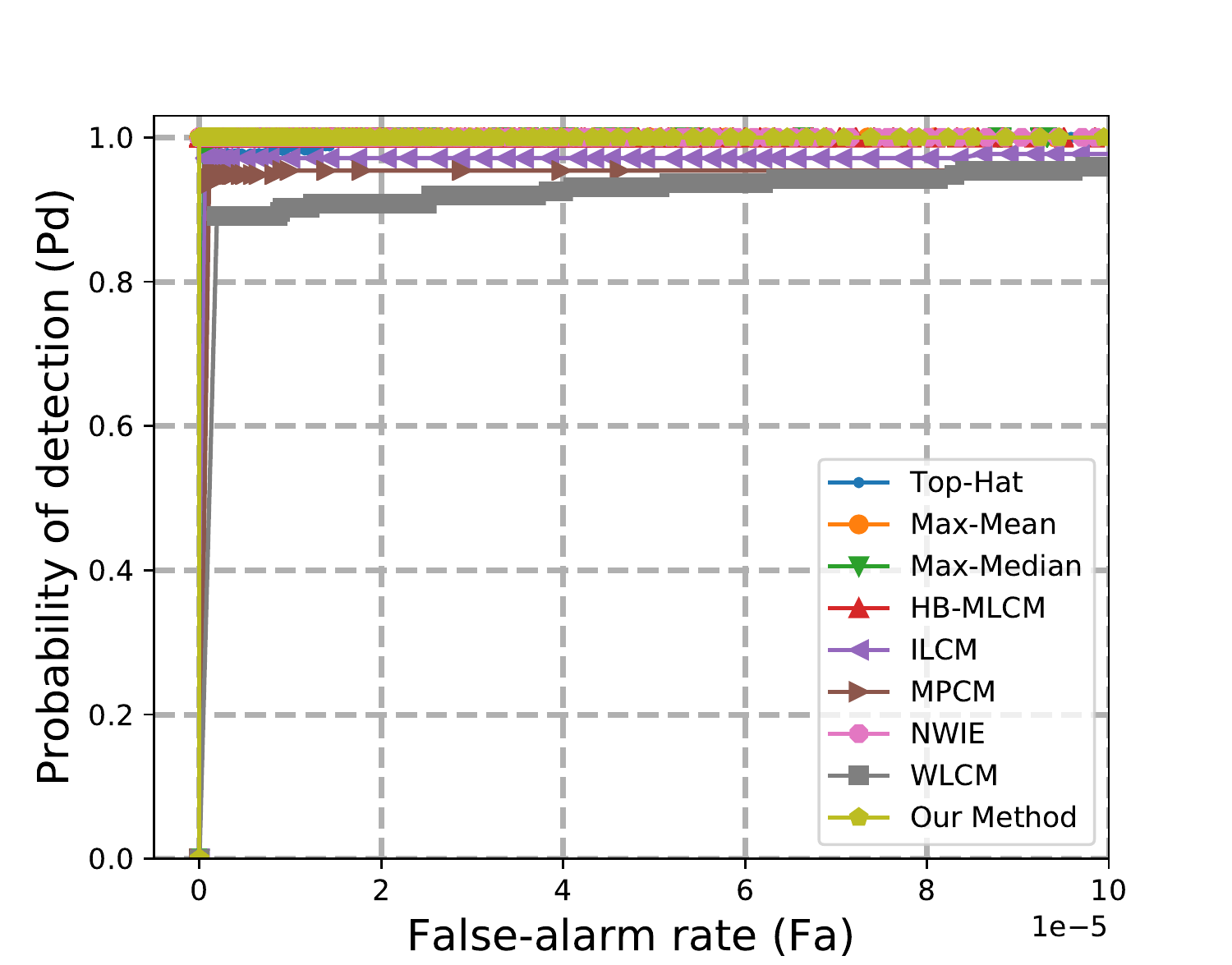}%
		\label{seq03_roc}}\\
	\subfloat[]{\includegraphics[width=0.33\textwidth]{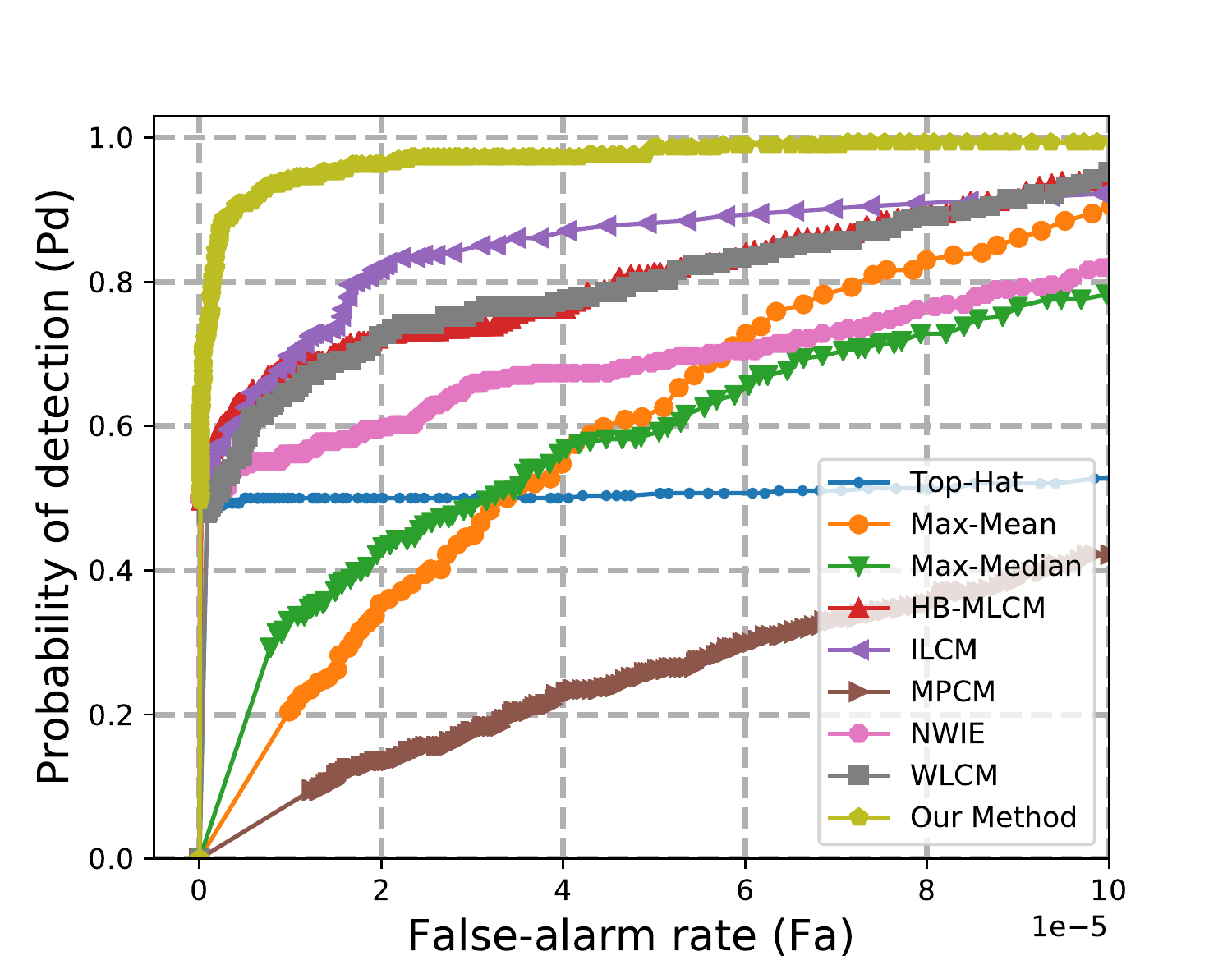}%
		\label{seq_roc04}}
	\hfil
	\subfloat[]{\includegraphics[width=0.33\textwidth]{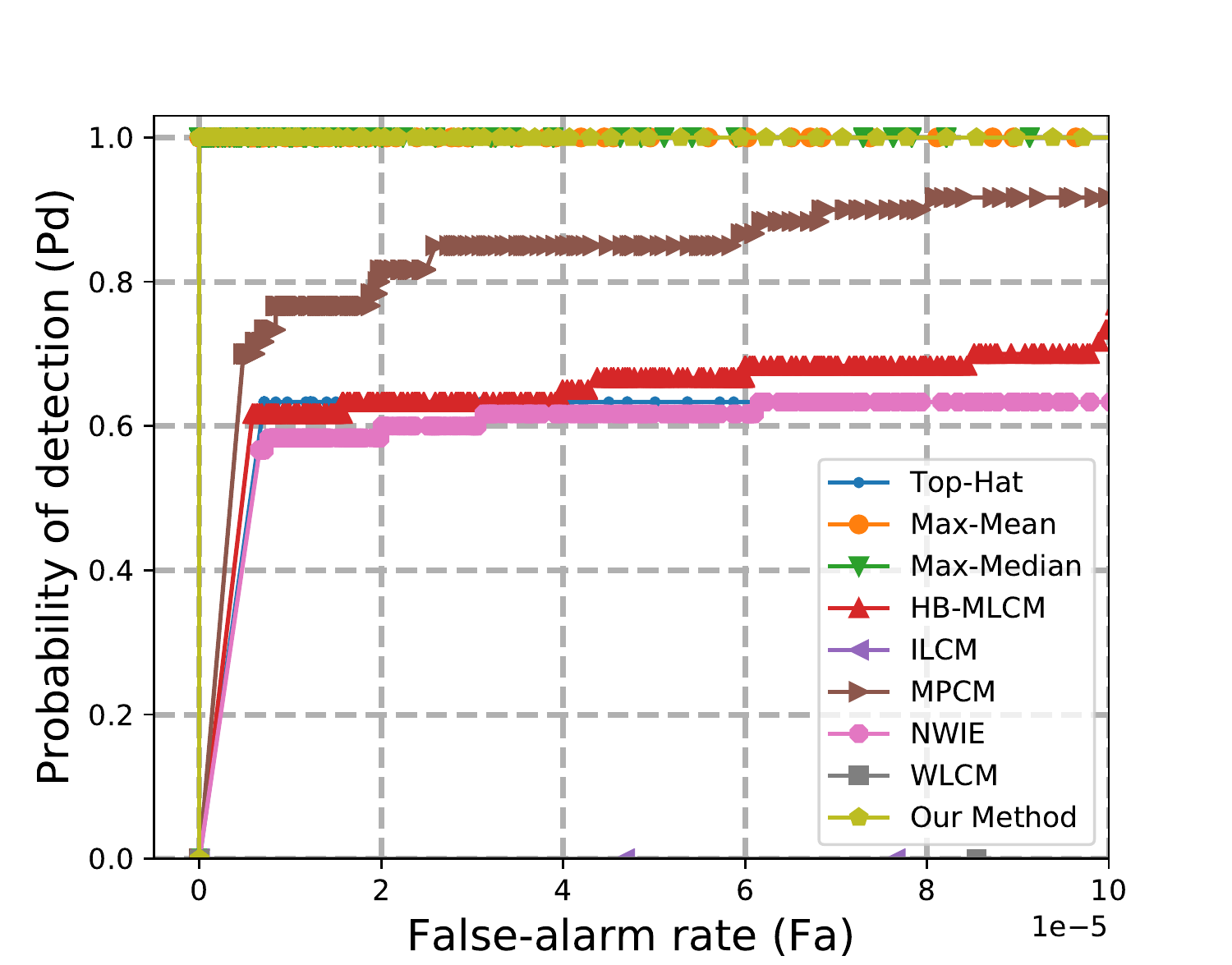}%
		\label{seq_roc05}}
	\hfil
	\subfloat[]{\includegraphics[width=0.33\textwidth]{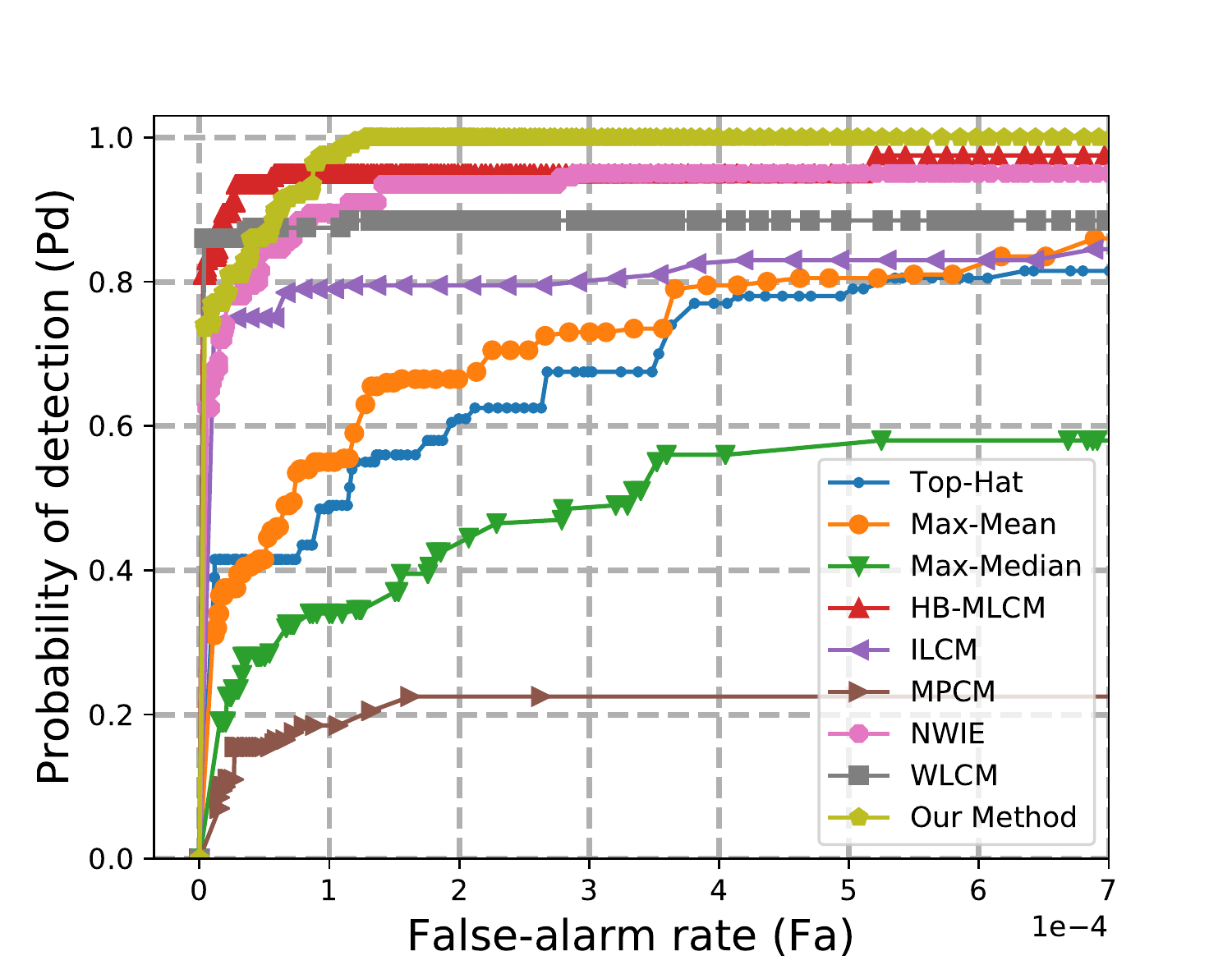}%
		\label{seq_roc06}}
	\caption{The receiver operating characteristic (ROC) curves of baseline methods and TBC-Net for six data sets. (a) Data set 1. (b) Data set 2. (c) Data set 3. (d) Data set 4 (e) Data set 5 (f) Data set 6}
	\label{ROC}
\end{figure*}

Figure \ref{ROC} shows the ROC curves achieved by different algorithms on 6 data sets. It can be seen that due to the better background suppression of TBC-Net, it can achieve a higher detection probability at a lower false alarm rate. Especially on data sets 1, 2, 4, and 6, where there are a lot of interference from branches, roads, and light noise in the background, TBC-Net can be significantly better than other detection methods. This also shows from one side that the TBC-Net trained on large-size images rather than local features can effectively suppress some complicated interference backgrounds better than the traditional methods because the network can not only learn the local features of the target but also captures wide range and high-level information from the entire image.

Table \ref{bsf} and \ref{scrg} show the BSF and SCRG for different algorithms, respectively. For both SCRG and BSF, the higher the value, the better the detection performance. 

Except for data set 5, TBC-Net has the highest BSF on other data. On data set 5, although the background generated by MPCM is quite flat, its BSF is higher than TBC-Net, but it incorrectly enhances the background. 

For SCRG, TBC-Net can achieve the best results on all data sets. Therefore, considering the results of comprehensive background suppression and target enhancement, the performance of TBC-Net is significantly better than other algorithms.

\begin{table*}[!h]
	\renewcommand{\arraystretch}{1.3}
	\centering
	\caption{BSF values of infrared data sets 1-6 processed by different methods}
	\label{bsf}
	\begin{tabular}{@{}cccccccccc@{}}
		\toprule
		& Top-Hat & Max-Mean & Max-Median & HB-MLCM & ILCM  & MPCM            & NWIE   & WLCM  & \textbf{Ours}            \\ \midrule
		Data set 1 & 2.221   & 4.794    & 5.162      & 28.134  & 0.909 & 9.377           & 22.302 & 1.210 & \textbf{32.754} \\
		Data set 2 & 2.495   & 6.006    & 7.756      & 13.656  & 0.976 & 23.984          & 8.254  & 1.211 & \textbf{24.276} \\
		Data set 3 & 1.477   & 2.389    & 3.344      & 19.305  & 0.980 & 2.676           & 8.162  & 1.707 & \textbf{43.812} \\
		Data set 4 & 1.644   & 3.338    & 5.479      & 7.196   & 1.013 & 9.442           & 6.903  & 1.091 & \textbf{10.080} \\
		Data set 5 & 3.443   & 9.139    & 12.095     & 35.511  & 1.011 & \textbf{52.126} & 25.124 & 1.482 & 40.293          \\
		Data set 6 & 1.453   & 2.614    & 4.050      & 19.026  & 1.097 & 2.907           & 6.703  & 1.592 & \textbf{45.827} \\ \bottomrule
	\end{tabular}
\end{table*}

\begin{table*}[!h]
	\renewcommand{\arraystretch}{1.3}
	\centering
	\caption{SCRG values of infrared data sets 1-6 processed by different methods}
	\label{scrg}
	\begin{tabular}{@{}cccccccccc@{}}
		\toprule
		& Top-Hat & Max-Mean & Max-Median & HB-MLCM & ILCM  & MPCM   & NWIE   & WLCM  & \textbf{Ours}     \\ \midrule
		Data set 1 & 1.189   & 0.291    & 0.364      & 2.150   & 0.323 & 2.898  & 2.530  & 0.202 & \textbf{42.914}   \\
		Data set 2 & 0.741   & 1.068    & 0.433      & 2.742   & 0.814 & 2.237  & 5.194  & 0.656 & \textbf{17.886}   \\
		Data set 3 & 1.578   & 0.719    & 0.288      & 6.230   & 1.336 & 1.504  & 5.699  & 0.581 & \textbf{113.542}  \\
		Data set 4 & 1.969   & 0.333    & 0.984      & 3.420   & 1.228 & 1.053  & 5.443  & 1.448 & \textbf{174.564}  \\
		Data set 5 & 8.734   & 7.858    & 10.829     & 62.391  & 7.509 & 25.957 & 79.268 & 3.082 & \textbf{1045.046} \\
		Data set 6 & 1.849   & 0.417    & 0.261      & 4.505   & 0.586 & 1.854  & 6.818  & 0.107 & \textbf{124.383}  \\ \bottomrule
	\end{tabular}
\end{table*}

\subsection{Performance Evaluation}

\subsubsection{Performance Comparison with Other Methods} 
The NVIDIA Jetson AGX Xavier development board is an embedded AI accelerator widely used in robotics and autonomous driving. In addition to running TBC-Net on the CPU and GPU, we deploy it on the Jetson AGX Xavier development board to evaluate its performance on mobile devices.

For baseline methods, we only count the execution time on the CPU. For TBC-Net, we count its execution time on the CPU, GPU, and AGX Xavier development board respectively. The test environment is that the PC has a 3.60-GHz Intel i9-9900k CPU and 64.0-GB memory, the GPU is NVIDIA RTX 2080Ti, and the development board is Jestson AGX Xavier. All traditional methods are implemented in MATLAB 2018b software, while the TBC-Net is implemented using Pytorch 1.0.1.


We first read all image data into memory, and only count the time from image input to processing completion to ignore data preparation time. Table \ref{time_consumption} shows the average single-frame processing time of different baseline algorithms on the CPU and the average single-frame processing time of TBC-Net on CPU/GPU/Jetson AGX Xavier tested on all sequences, and we use C, G, B to distinguish TBC-Net execution time on CPU, GPU, and development board. The Jetson AGX Xavier supports 10W, 15W, 30W three power modes to suit different applications. We use the 30W power mode here.

\begin{table*}[!h]
	\renewcommand{\arraystretch}{1.3}
	\centering
	\caption{Time consumption of different algorithms.}
	\label{time_consumption}
	\begin{tabular}{|c|c|c|c|c|c|c|c|c|c|c|c|}
		\hline
		Method   & Top-Hat & Max-Mean & Max-Median & HB-MLCM & ILCM  & MPCM  & NWIE & WLCM & \textbf{Ours (C)} & \textbf{Ours (G)} & \textbf{Ours (B)} \\ \hline
		Time (s) & 0.0086  & 1.021    & 1.010      & 0.010   & 0.039 & 0.028 & 0.11 & 0.15 & \textbf{0.051}    & \textbf{0.002}    & \textbf{0.014}    \\ \hline
	\end{tabular}
\end{table*}

The authors of the above methods did not disclose their source code,  so we reproduce these algorithms according to original papers. The execution efficiency of some algorithms may not be met or is inconsistent with the efficiency reported in the original paper.

In terms of CPU running time, although TBC-Net is slower than algorithms such as Top-Hat, HB-MLCM, ILCM, and MPCM, its detection performance is still significantly better than these algorithms. Moreover, from a more practical perspective, TBC-Net runs efficiently on embedded devices with neural network acceleration hardware, while other traditional algorithms do not have as good parallel optimization space on embedded devices to improve performance.

\subsubsection{Performance Comparison under Different Power Modes}

Since most deep learning accelerators are designed for large data throughput operations, processing data in batches will be faster than single-frame processing, so we also test the average single-frame processing time for TBC-Net at different batch sizes. The significance of batch processing is that in some embedded scenarios, an accelerator may be required to process video images acquired by multiple sensors simultaneously. The test results are shown in Table \ref{inference}. In the case of batch-frame processing, TBC-Net can achieve frame rates of up to 50, 91, and 175 in 10W, 15W, and 30W power mode, respectively, indicating that it is suitable for fast target detection in embedded low-power scenarios.

\begin{table*}[!h]
\renewcommand{\arraystretch}{1.3}
\centering
\caption{TBC-Net inference time with different power mode and image batch size.}
\label{inference}
\begin{tabular}{c|c|c|c|c|c|c|c}
\hline
\multicolumn{2}{c|}{}                                                                         & \multicolumn{6}{c}{Image batch size}                                                                      \\ \cline{3-8} 
\multicolumn{2}{c|}{\multirow{-2}{*}{\begin{tabular}[c]{@{}c@{}}Time (s)\end{tabular}}} & 1                            & 2      & 4      & 8      & 16                            & 32     \\ \hline
                                                                             & 10W            & 0.036                        & 0.029  & 0.027  & 0.021  & 0.020                         & 0.022  \\ \cline{2-8} 
                                                                             & 15W            & 0.027                        & 0.017  & 0.016  & 0.012  & 0.011                         & 0.011  \\ \cline{2-8} 
\multirow{-3}{*}{\begin{tabular}[c]{@{}c@{}}Power\\ mode\end{tabular}}            & 30W            & \textbf{0.014} & 0.0079 & 0.0065 & 0.0059 & \textbf{0.0057} & 0.0059 \\ \hline
\end{tabular}
\end{table*}

\section{Conclusion}

In this paper, we presented a novel lightweight convolutional neural network TBC-Net, including a target extraction module and a semantic constraint module for infrared small target detection. With the help of semantic constraint module, joint loss function, synthesis data, and corresponding training method, we solved the problem caused by the extreme imbalance of targets and the background when using CNN to learn small target features on images of $256\times 256$ resolution. The input image is processed by the target extraction module to directly obtain the target image, thereby achieving end-to-end small target detection. Moreover, through storage space and computational complexity analysis, TBC-Net is well suited for deployment in embedded systems that support neural network acceleration. The experimental results show that TBC-Net trained from large-size images can better suppress the complex interference in the background compared to the detection algorithms using only local features. Besides, TBC-Net realizes real-time detection on the NVIDIA Jetson AGX Xavier development board. These advantages make TBC-Net have great application potential in applications such as infrared detection and search using drones.

\ifCLASSOPTIONcaptionsoff
  \newpage
\fi

\bibliographystyle{IEEEtran}
\bibliography{IEEEabrv,bib/paper}

\end{document}